\PassOptionsToPackage{dvipsnames}{xcolor}
\IfFileExists{awomo.cls}{
    \documentclass[onecolumn]{awomo}
}{
    \documentclass[onecolumn]{./awomo}
}

\makeatletter
\def\input@path{{./}{./}}
\makeatother
\graphicspath{{./}{./}}

\usepackage{arydshln}
\usepackage{color}
\usepackage{xcolor}
\usepackage{colortbl}
\definecolor{lightgray}{rgb}{0.83, 0.83, 0.83}

\makeatletter
\providecommand*{\theHALG@line}{\thealgorithm.\arabic{ALG@line}}
\makeatother

\title{StrucPhysVideo: Learning Physical Dynamics from Structured Captions and Robot Actions\\[4pt]\Large}

\author{
\begin{center}
    Awomo-WM Team\footnotemark[1]
\end{center}
}

\begin{document}

\abstract{
Modeling physical dynamics---how objects move, interact, and change state---is central to video world models for embodied AI. We present \textbf{StrucPhysVideo}, a family of video world models that bridges physics-focused data curation with language- and action-conditioned prediction of scene evolution. Our data pipeline combines motion-aware video segmentation, quality and content filtering, and physical relevance verification with structured annotations of objects, materials, and temporally localized interactions. By disentangling camera motion from object behavior and explicitly describing contact, deformation, and state transitions, the pipeline provides supervision grounded in observable physical events.
Building on these data, we introduce StrucPhysVideo-TI2V, a sparse Mixture-of-Experts (MoE) text-image-to-video model trained with a curriculum that progressively emphasizes physical dynamics while retaining general-domain video data. StrucPhysVideo-TI2V achieves state-of-the-art performance on \textit{Physics-IQ Verified}, scoring \textbf{45.5\%} and outperforming Cosmos3-Super-Image2Video by 2.8 percentage points. Caption ablations across backbones further demonstrate the effectiveness of physics-focused supervision. We further extend StrucPhysVideo-TI2V to StrucPhysVideo-IA2V, an interactive image-action-to-video world model that predicts visual outcomes from robot end-effector commands. Action conditioning, causal autoregressive generation, and few-step distillation enable incremental robot rollouts with only four denoising steps. Together, StrucPhysVideo advances physical dynamics modeling from image- and language-conditioned video prediction toward action-driven interaction.

\textbf{Website}: \href{https://westlakedi-awomo.github.io/StrucPhysVideo-Page/}{https://westlakedi-awomo.github.io/StrucPhysVideo-Page/}\\
\textbf{Github}:  \href{https://github.com/westlakedi-awomo/StrucPhysVideo}{https://github.com/westlakedi-awomo/StrucPhysVideo}
}

\maketitle
\footnotetext[1]{See Contributions section for full author list. Please send correspondence to \href{mailto:info@westlakedi.com}{info@westlakedi.com}.}

\begin{figure}[!ht]
  \centering
  \includegraphics[width=\linewidth]{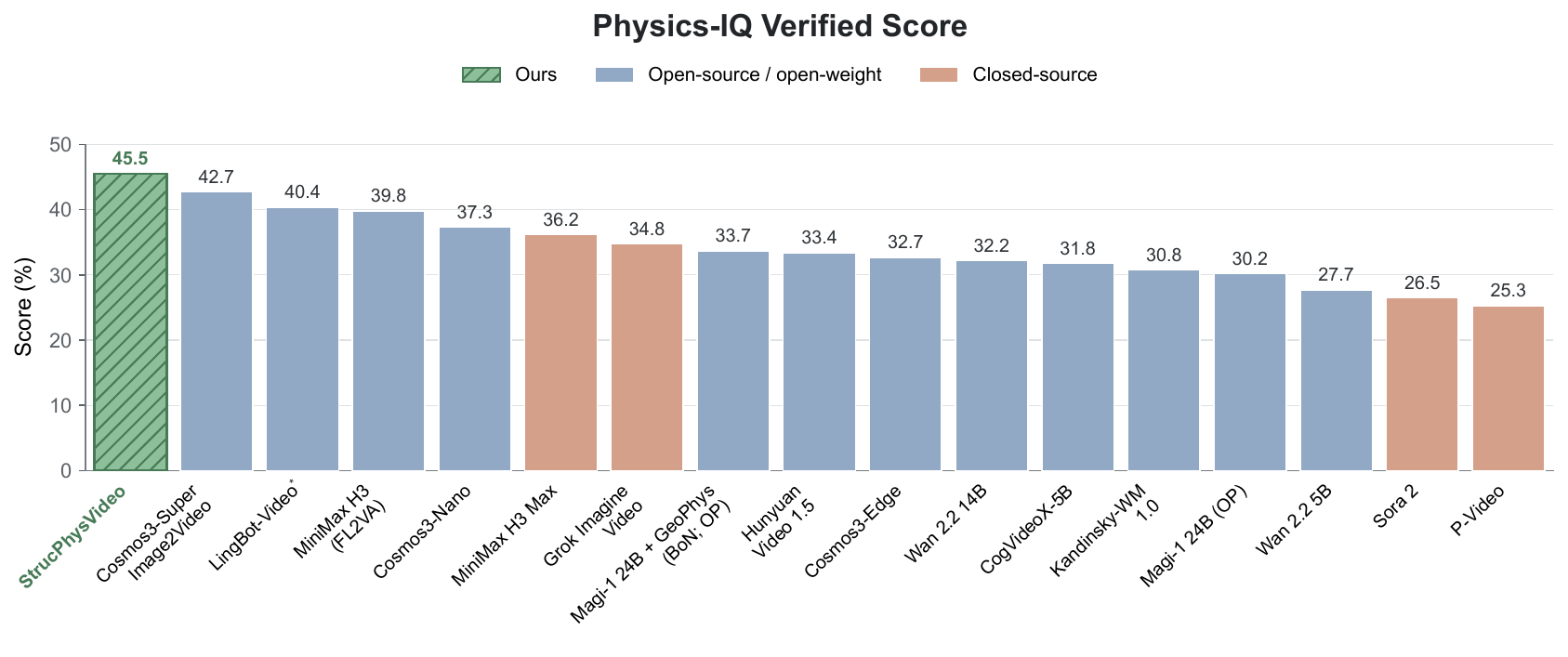}
\caption{Physics-IQ Verified benchmark Results~\cite{radsch2026physicsiqverified}. StrucPhysVideo achieves the highest Physics-IQ Verified score among all compared models in the text-image-to-video setting. * denotes our reproduction, and results for all other models are taken from the benchmark snapshot dated 16 September 2026~\cite{physicsiqleaderboard2026}.}
  \label{fig:physics_iq_ti2v}
\end{figure}

\justifying
\section{Introduction}
\label{sec:introduction}
Physical AI requires models that connect visual observations with the consequences of actions. Such models should capture how objects persist, how interactions alter their states, and how scenes evolve in response to agent actions. Video provides natural observations of these processes over time, making it a valuable medium for learning representations of the physical world. Developing such video models requires training data that preserve meaningful physical interactions, annotations grounded in physical events, and modeling frameworks that capture how actions drive scene evolution.

Preparing such data requires attention to the visual evidence available across time. Consider a video in which a gripper approaches a cup, grasps it, and lifts it from a table. A cut made during the grasp can separate the approach from the lift, leaving neither clip with the full progression. An editing transition may instead connect the cup on the table to the cup already in the air, without showing how it moved between these states. Poor exposure or persistent blur may also hide the contact point, even when the clip contains the full interaction. Filtering also requires care: a label printed on a bottle belongs to the physical scene, while an overlaid banner may obscure the interaction. Rejecting all clips with detected text would remove both. These cases motivate curation that considers temporal continuity, visual quality, and the context of the content being filtered.

\begin{figure}[!t]
    \centering
    \includegraphics[width=0.99\linewidth]{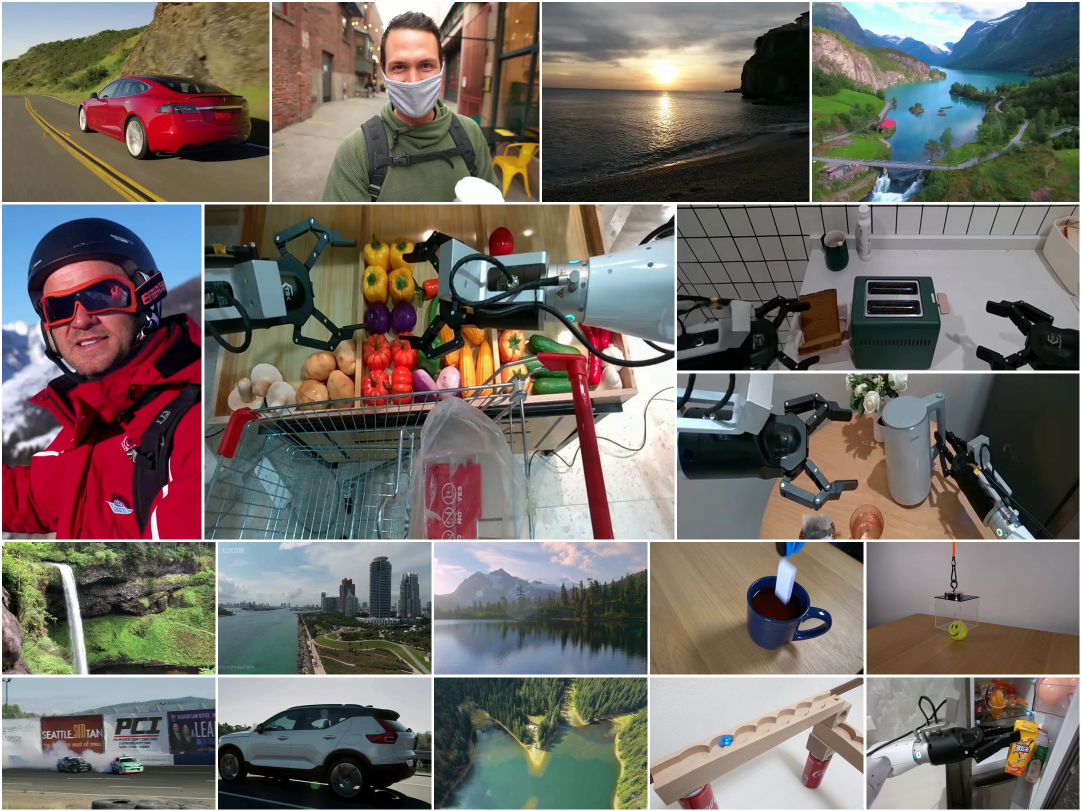}
    \caption{Representative StrucPhysVideo generations under both the text-image-to-video (TI2V) and image-action-to-video (IA2V) settings, spanning general-domain, physical-world, and embodied scenarios.}
    \label{fig:teaser}
\end{figure}

Language introduces another difficulty. A caption such as ``a robot handles a cup'' describes the general activity but does not specify which cup is involved, whether it is pushed or lifted, or how its state changes. In a scene with several objects, these details are necessary to ground an instruction such as ``Lift the red cup from the table.'' The order of approaching, grasping, and lifting also matters because each stage describes a different part of the interaction. Camera movement adds further ambiguity: a stationary cup can move across the image as the camera pans. Language supervision, therefore, needs to identify the participating objects, describe their actions in temporal order, and distinguish object behavior from camera motion. When contact is hidden by a hand or missed by frame sampling, the description should remain within the available visual evidence. 

Even with a clear instruction, the execution of an interaction can remain ambiguous. For example, ``Move the cup to the other side of the table'' could involve sliding it along the surface or lifting it, carrying it, and placing it down. These executions differ in their contact sequence and intermediate states, although they may produce the same final position. Action information provides an additional condition for guiding how the scene unfolds. If the supplied actions correspond to a lift, a generated sequence in which the cup continues to slide would be inconsistent with those actions. For embodied AI, the usefulness of the generated video depends on whether these visual changes follow the supplied interaction throughout the sequence.

In this work, we present \textbf{StrucPhysVideo}, a Physical AI model that connects the curation of our privately collected video data with language- and action-conditioned generation. We develop a cleaning and curation pipeline to prepare clips with observable interactions, and an language annotation pipeline that organizes scene context, entities, and temporal behaviors into structured descriptions. These descriptions provide a basis for constructing instructions grounded in the video content. We further add action information to guide generation, with the aim of maintaining consistency between the instruction, the supplied actions, and the resulting visual changes. Representative generations across general-domain, physical-world, and embodied scenarios are shown in Figure~\ref{fig:teaser}. Together, these components connect interaction-focused data curation and structured language supervision with language- and action-conditioned video generation. As a result, StrucPhysVideo achieves strong physical dynamics prediction performance, scoring 45.5\% on Physics-IQ Verified and ranking first among the 17 models in our comparison based on the 16 September 2026 benchmark snapshot, as shown in Figure~\ref{fig:physics_iq_ti2v}.

We make three contributions:
\begin{itemize}
    \item \textbf{Video cleaning and curation.} We propose a pipeline that combines shot detection, motion-aware segmentation, technical quality checks, and content filtering with targeted multimodal review to retain useful observations of physical interactions.
    \item \textbf{A structured language annotation pipeline.} We construct structured descriptions of scenes, entities, and temporally localized actions, separating camera motion from object behavior to produce instructions that are faithfully grounded in the video content.
    \item \textbf{Action-conditioned video generation.} We add action information to guide generation in StrucPhysVideo, with the objective of producing interactions consistent with the supplied actions while preserving visual quality and temporal continuity.
\end{itemize}
The remainder of this report is organized as follows. Section~\ref{sec:method} describes the model architecture and post-training procedures, including action-conditioned robot video generation. Section~\ref{sec:data_pipeline} details the data sourcing and curation pipeline, structured physical captioning, and physical taxonomy. Section~\ref{sec:experiments} presents evaluations of physical dynamics prediction, caption ablations, and controlled robot-video comparisons. Finally, Section~\ref{sec:conclusion} discusses limitations and directions for future research.

\section{Method}
\label{sec:method}


\subsection{StrucPhysVideo-TI2V}
\label{sec:strucphysvideo-ti2v}

StrucPhysVideo-TI2V is a latent generative model that predicts how an observed scene evolves under an image and language condition. Given a reference image $I_0$ and a caption $c$, it models
\begin{equation}
  p_\theta(I_{1:T}\mid I_0,c),
  \label{eq:ti2v_objective}
\end{equation}
while preserving the subject identity, spatial layout, camera geometry, and visual style established by $I_0$. The model treats the reference image as both a semantic condition and a boundary condition on the generated trajectory. This dual role directs model capacity toward learning plausible motion and state changes rather than reconstructing already observed visual information.

\begin{figure*}[!t]
  \centering
  \includegraphics[width=\textwidth]{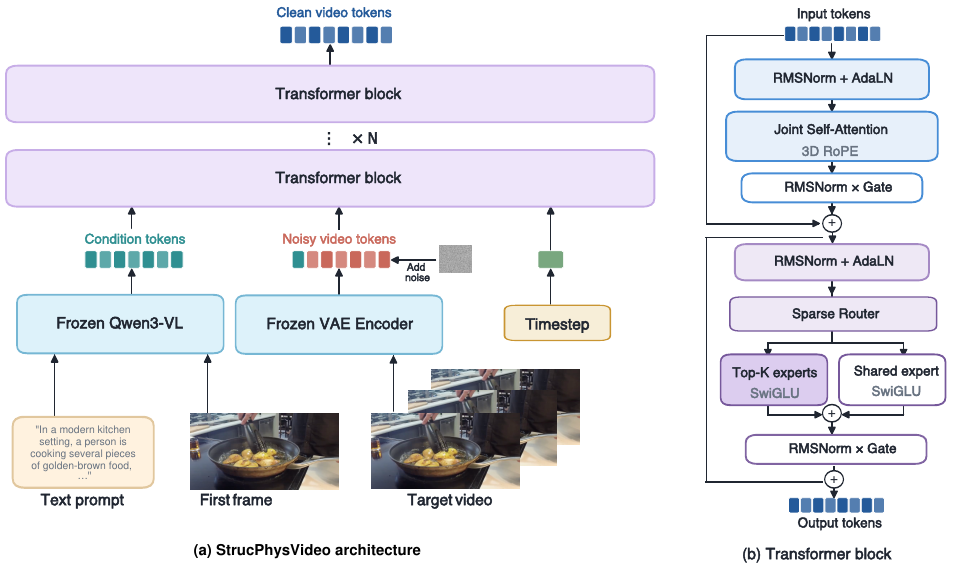}
  \caption{StrucPhysVideo-TI2V overall architecture. The frozen Qwen3-VL-32B encoder jointly encodes the text prompt and reference image, while the frozen VAE encodes the target video. The first clean video latent is retained as a reference and the remaining latents are corrupted along the flow-matching path. Multimodal condition tokens and video tokens interact through joint self-attention in the StrucPhysVideo Transformer. Each block combines timestep-modulated attention with routed and shared experts, and the model predicts the latent velocity field used to supervise future-frame generation.}
  \label{fig:strucphysvideo-overview}
\end{figure*}

Figure~\ref{fig:strucphysvideo-overview} summarizes the overall architecture. A frozen Qwen3-VL-32B encoder~\cite{bai2025qwen3vl} jointly processes $c$ and $I_0$ to produce a multimodal condition $\mathbf{h}_c=\mathcal{Q}(c,I_0)$. In parallel, a frozen Wan video variational autoencoder (VAE)~\cite{wan2025} maps the target video $\mathbf{V}=\{I_0,I_1,\ldots,I_T\}$ to a normalized clean latent $\mathbf{z}_0=\mathcal{N}(\mathcal{E}(\mathbf{V}))$. The trainable StrucPhysVideo backbone receives the multimodal condition, a corrupted video latent, and the flow timestep, and predicts a velocity field in latent space. The VAE decoder $\mathcal{D}$ maps the final latent trajectory back to RGB video.

\subsubsection{Dual-Path Reference Conditioning}
\label{sec:awomo-conditioning}

The semantic path uses the final hidden states of Qwen3-VL-32B rather than generating an intermediate rewritten caption. Because the encoder observes both modalities, $\mathbf{h}_c$ captures visible entities and appearance together with linguistic descriptions of motion, temporal order, and camera behavior. A learned projection maps this condition into the backbone feature space. The latent path supplies a complementary low-level constraint. During training, the reference latent $\mathbf{z}_0^{\mathrm{ref}}$ is taken from the first temporal position of the VAE encoding of the complete target video, avoiding a mismatch between independently encoded images and the temporal boundary behavior of the video VAE.

After spatiotemporal patchification, video and condition tokens are concatenated and processed by non-causal joint self-attention. Unlike a design that injects the condition through occasional cross-attention layers, this formulation allows language, reference-image features, and evolving video features to interact in every Transformer block. Three-axis rotary position embeddings, adapted from RoPE~\cite{su2021roformer}, distinguish temporal and spatial locations for video tokens while retaining an ordered position convention for condition tokens.

\subsubsection{Sparse Spatiotemporal Transformer}
\label{sec:awomo-transformer}

The trainable backbone is an approximately 30-billion-parameter spatiotemporal Transformer with 48 blocks, following the sparse video-pretraining design of LingBot-Video~\cite{ma2026lingbotvideo}. Every feed-forward sublayer uses a sparse mixture-of-experts (MoE) module containing routed experts and an always-active shared expert. For a token representation $\mathbf{x}$, the router selects a restricted top-$k$ set $\mathcal{T}(\mathbf{x})$, and the MoE output is
\begin{equation}
  \operatorname{MoE}(\mathbf{x})=
  \sum_{i\in\mathcal{T}(\mathbf{x})}\alpha_i E_i(\mathbf{x})
  +E_{\mathrm{shared}}(\mathbf{x}),
  \label{eq:awomo-moe}
\end{equation}
where $E_i$ denotes a routed expert and $\alpha_i$ is its normalized routing weight. Group-constrained routing limits the candidate experts before top-$k$ selection. This architecture increases model capacity while keeping the amount of expert computation per token bounded. Timestep-conditioned adaptive normalization, following the conditioning strategy used in diffusion Transformers~\cite{peebles2023dit}, and gated residual connections modulate both the attention and MoE branches across the flow trajectory.

\subsubsection{First-Frame-Constrained Flow Matching}
\label{sec:awomo-objective}

We use the flow-matching formulation~\cite{lipman2023flow} and the same notation as the action-conditioned model in Section~\ref{sec:action_world_model}. For a clean video latent $\mathbf{z}_0$, Gaussian noise $\boldsymbol{\epsilon}$, and flow time $\tau$, the corrupted state and target velocity are
\begin{equation}
  \mathbf{z}_\tau=(1-\tau)\mathbf{z}_0+\tau\boldsymbol{\epsilon},
  \qquad \mathbf{v}^{\star}=\boldsymbol{\epsilon}-\mathbf{z}_0.
  \label{eq:awomo-flow-path}
\end{equation}
Before each denoiser evaluation, the first temporal position of $\mathbf{z}_\tau$ is replaced by $\mathbf{z}_0^{\mathrm{ref}}$. The model therefore observes the clean initial state while predicting $\mathbf{v}_\theta(\mathbf{z}_\tau,I_0,c)$ for the full latent sequence. The reference position is excluded from the regression loss:
\begin{equation}
  \mathcal{L}_{\mathrm{FM}}=
  \mathbb{E}_{\tau,\boldsymbol{\epsilon}}
  \left[
  w(\tau)\left\|
  \mathbf{M}\odot\left(\mathbf{v}_\theta-\mathbf{v}^{\star}\right)
  \right\|_2^2
  \right],
  \label{eq:awomo-flow-loss}
\end{equation}
where $\mathbf{M}$ masks the reference position and retains all future positions. Thus, the initial latent participates in contextual reasoning but is not itself a denoising target. Appendix~\ref{app:ti2v-algorithms}, Algorithm~\ref{alg:awomo-training}, gives the corresponding single-step training procedure.

\subsubsection{Inference}
\label{sec:awomo-inference}

At inference, Qwen3-VL-32B again encodes $(c,I_0)$, and the VAE image-encoding path produces $\mathbf{z}_0^{\mathrm{ref}}$. Sampling begins from a Gaussian video latent whose first temporal position is replaced by this clean reference. A Flow-UniPC solver, adapted from the UniPC predictor-corrector framework~\cite{zhao2023unipc}, integrates the predicted velocity field, and the reference position is re-clamped after every solver update to prevent numerical drift. When classifier-free guidance~\cite{ho2022cfg} is used, conditional and negative-condition predictions are combined as
\begin{equation}
  \widehat{\mathbf{v}}=
  \mathbf{v}_{\mathrm{uncond}}+
  s_t\left(\mathbf{v}_{\mathrm{cond}}-\mathbf{v}_{\mathrm{uncond}}\right).
  \label{eq:awomo-cfg}
\end{equation}
The final latent is inverse-normalized and decoded by $\mathcal{D}$. Appendix~\ref{app:ti2v-algorithms}, Algorithm~\ref{alg:awomo-inference}, provides the full inference procedure. Concrete training resolutions, frame counts, batch sizes, optimization settings, precision choices, and distributed execution details are reported in Section~\ref{sec:ti2v_implementation}.




\subsection{StrucPhysVideo-IA2V}
\label{sec:action_world_model}

We further extend StrucPhysVideo-TI2V into an interactive robot world model, \textbf{StrucPhysVideo-IA2V}, that predicts how visual scenes evolve in response to robot command trajectories. To enable real-time interaction, we convert the bidirectional video DiT into a causal few-step generator through a three-phase fine-tuning pipeline.

\subsubsection{Problem Formulation and System Design}
\label{sec:robot_problem}

Let $I_0$ be the first RGB observation and let $\mathbf{a}_{0:K-1}$ be a sequence of robot actions. Our image-and-action-to-video (IA2V) model represents
\begin{equation}
  p_\theta(I_{1:T}\mid I_0,\mathbf{a}_{0:K-1}),
  \label{eq:ia2v_objective}
\end{equation}
while the text-image-action-to-video (TIA2V) variant additionally conditions on a textual command $c$. Both variants use the same visual backbone and action interface; IA2V fixes the text context to the embedding of the empty string.

The system is organized into three phases. First, action injection adapts a pretrained bidirectional backbone to follow end-effector commands over a fixed video window. Second, teacher forcing AR Diffusion training and Causal ODE converts the resulting model into a streaming autoregressive generator. Finally, asymmetric DMD performs distribution-level refinement by aligning the autoregressive generator with the high-quality bidirectional teacher, recovering generation quality lost during causalization.

\subsubsection{Robot Data and Action Representation}
\label{sec:ee20_representation}

Each sample contains a $960\times736$ head-camera clip and the commanded dual-arm EEF trajectory, sampled at 5 FPS and 15 Hz, respectively. For arm $r\in\{L,R\}$, the input vector for the model is
\begin{equation}
  \mathbf{a}^{r}_k = [\,\mathbf{p}^{r}_k,\,\rho(\mathbf{q}^{r}_k),\,g^{r}_k\,]\in\mathbb{R}^{10},
\end{equation}
where $\mathbf{p}\in\mathbb{R}^3$ is Cartesian position, $\mathbf{q}$ is an $xyzw$ quaternion, $g$ is the gripper target, and $\rho(\mathbf{q})\in\mathbb{R}^{6}$ contains the first two columns of the corresponding rotation matrix. This continuous 6D representation avoids the quaternion sign ambiguity and is well suited to regression in neural networks~\cite{zhou2019rotation}. Concatenating the left and right arms yields
\begin{equation}
  \mathbf{a}_k=[\mathbf{a}^{L}_k,\mathbf{a}^{R}_k]\in\mathbb{R}^{20}.
\end{equation}
Every coordinate is normalized with corpus-level mean and standard deviation.

\paragraph{Frame-aligned action windows.}
At 15 Hz, every three action commands correspond to the transition between two adjacent video frames sampled at 5 fps. Thus, commands $[3i,3i+3)$ correspond to the transition from video frame $i$ to frame $i+1$. Since each video latent represents four consecutive video frames, it is aligned with a window of twelve action commands.

\begin{figure*}[t]
  \centering
  \includegraphics[width=0.82\textwidth]{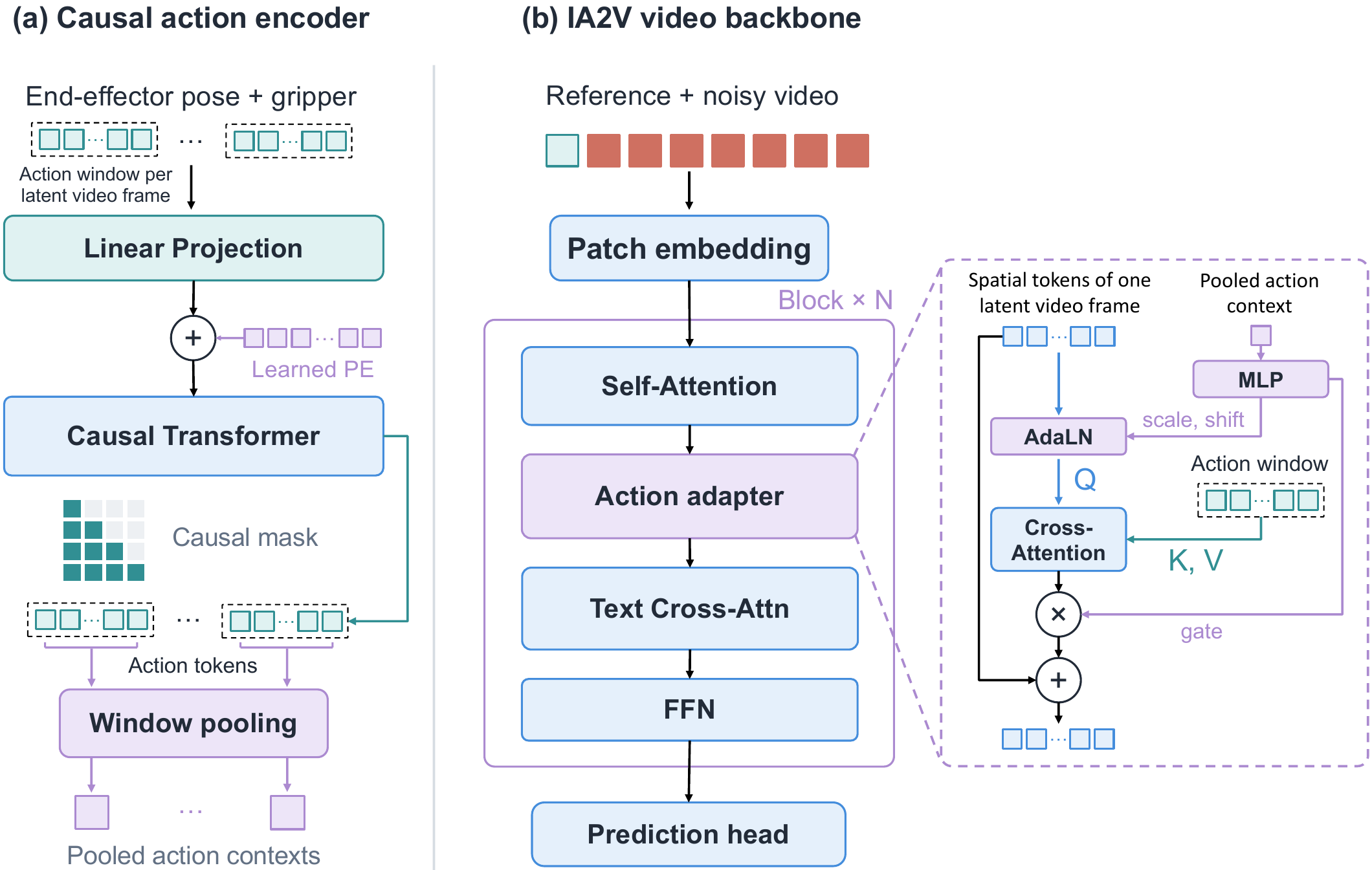}
  \caption{StrucPhysVideo-IA2V architecture. A causal action encoder maps raw action commands into latent action tokens and pooled contexts aligned with each video latent frame. In each DiT block, the pooled context modulates the video queries through AdaLN, while the corresponding action window provides the keys and values for cross-attention. A gated residual integrates action-conditioned features into the video stream.}
  \label{fig:awomo-ia2v}
\end{figure*}

\paragraph{Causal action encoder.}
As shown in Figure~\ref{fig:awomo-ia2v}, the normalized action trajectory $\mathbf{a}_{0:K-1}$ is first projected into a 512-dimensional embedding space, followed by a four-layer Transformer encoder with a strict causal attention mask:
\begin{equation}
\mathbf{u}_{0:K-1}=E_{\mathrm{act}}(\operatorname{LN}(\mathbf{a}_{0:K-1})W_{\mathrm{in}}+\mathbf{p}_{0:K-1}),
\qquad \mathbf{u}_k\ \text{depends only on}\ \mathbf{a}_{0:k}.
\end{equation}
Here, $\mathbf{p}_k$ denotes the learned positional embedding for action step $k$. The encoder employs a strict upper-triangular causal attention mask, ensuring that each action token attends only to past and current commands. Finally, attention pooling is performed within each action window to produce a frame-level action context $\mathbf{h}_f$, allowing each video latent to condition exclusively on the action commands associated with its corresponding temporal interval.

\subsubsection{Bidirectional IA2V Training}
\label{sec:bidirectional_robot_training}

\paragraph{Reference-image conditioning.}
For a clean latent video $\mathbf{z}_0$ and Gaussian noise $\boldsymbol{\epsilon}$, rectified-flow corruption constructs
\begin{equation}
  \mathbf{z}_\tau=(1-\tau)\mathbf{z}_0+\tau\boldsymbol{\epsilon},
  \qquad \mathbf{v}^{\star}=\boldsymbol{\epsilon}-\mathbf{z}_0,
\end{equation}
and trains the network to predict the velocity field~\cite{lipman2023flow}. We replace the first noisy latent by the clean reference latent before every denoiser call and exclude it from the loss:
\begin{equation}
  \mathcal{L}_{\mathrm{FM}}=
  \mathbb{E}_{\tau,\boldsymbol{\epsilon}}
  \left[\frac{1}{N}\sum_{f=1}^{N}w(\tau)
  \left\|\mathbf{v}_\theta(\mathbf{z}_\tau,I_0,\mathbf{a},c)_f-
  \mathbf{v}^{\star}_f\right\|_2^2\right].
  \label{eq:robot_flow_loss}
\end{equation}
We use token-wise diffusion time-steps; reference tokens are marked with $\tau=0$, while generated tokens use the sampled timestep. At inference, the solver updates the whole latent tensor, so the reference latent is re-clamped after every solver step. This preserves the initial image exactly in latent space.

\paragraph{Per-block AdaLN cross-attention adapter.}
As illustrated in Figure~\ref{fig:awomo-ia2v}, we inject action control after video self-attention and before text cross-attention in every DiT block. For the video tokens $\mathbf{x}_f$ belonging to latent frame $f$, a low-rank modulation network predicts shift, scale, and gate vectors from $\mathbf{h}_f$:
\begin{align}
  (\Delta\boldsymbol{\mu}_f,\Delta\boldsymbol{\sigma}_f,\mathbf{g}_f)
  &=M_\ell(\mathbf{h}_f),\\
  \widetilde{\mathbf{x}}_f
  &=(1+\Delta\boldsymbol{\sigma}_f)\odot\operatorname{LN}(\mathbf{x}_f)
    +\Delta\boldsymbol{\mu}_f,\\
  \mathbf{r}_f
  &=W_o\operatorname{Attn}\!\left(
    W_q\widetilde{\mathbf{x}}_f,
    W_k\mathbf{u}_{\mathcal{W}_f},
    W_v\mathbf{u}_{\mathcal{W}_f}\right),\\
  \mathbf{x}_f&\leftarrow\mathbf{x}_f+\tanh(\mathbf{g}_f)\odot\mathbf{r}_f.
  \label{eq:action_adapter}
\end{align}
Specifically, AdaLN first modulates the visual queries according to the corresponding trajectory. Cross-attention then retrieves fine-grained action tokens as keys and values, and the resulting action features are injected into the video stream through a gated residual connection. The gate is zero-initialized to preserve the functionality of the pretrained backbone at initialization. The latent-zero frame serves as the observed reference and therefore receives no action residual.
During training, we use a learning rate of $2\times10^{-6}$ for the pretrained backbone and a $25\times$ larger learning rate for the newly initialized action modules.

\paragraph{Unconditional Dropout for Teacher Guidance.}
During bidirectional action-injection training, we apply unconditional dropout to enable action CFG~\cite{ho2022cfg} in the teacher used for subsequent distillation. With probability $p_a$, the complete action sequence is replaced by a \textit{learned null-action token}. On the mixed general-domain T2V/TI2V data, text dropout replaces the complete text context with the empty-prompt context with probability $p_t$; IA2V samples already use empty text. The text and action dropout masks are sampled independently where both conditions are present. This trains the conditional and null-condition branches needed to construct a teacher signal with stronger action guidance.
Let $C$ denote supplied text plus action, $N$ negative text plus action, $T$ supplied text plus null action, and $U$ negative text plus null action. Teacher-side guidance supports
\begin{align}
  \widehat{\mathbf{v}}_{\mathrm{text}}
  &=N+s_t(C-N),\label{eq:text_cfg}\\
  \widehat{\mathbf{v}}_{\mathrm{action}}
  &=T+s_a(C-T),\label{eq:action_cfg}\\
  \widehat{\mathbf{v}}_{\mathrm{joint}}
  &=U+s_t(T-U)+s_a(C-T).\label{eq:joint_cfg}
\end{align}
These guidance scales describe the teacher, not the deployed student. Subsequent distillation, as described in the next section, transfers the teacher's guidance into a single-branch causal few-step generator.

\subsubsection{Causalization and Few-Step Distillation}
\label{sec:robot_causal_dmd}

The bidirectional model provides a high-quality, action-controllable teacher but cannot stream efficiently. Following Causal Forcing and Causal Forcing++~\cite{zhu2026causalforcing,zhao2026causalforcingpp}, we convert it into a few-step autoregressive model in three stages. The same first-image and EE20 action conditions are supplied throughout all stages.

\paragraph{Stage 1: autoregressive diffusion training.}
We initialize a causal model from the bidirectional checkpoint, partition the latent sequence into temporal chunks, and replace full temporal attention by a causal mask. For chunk $i$, clean ground-truth history $\mathbf{z}^{<i}_{\mathrm{gt}}$ is concatenated with a noised current chunk $\mathbf{z}^{i}_\tau$. Teacher-forced flow matching trains
\begin{equation}
  \mathbf{v}_{\mathrm{AR}}(\mathbf{z}^{i}_\tau,
  \mathbf{z}^{<i}_{\mathrm{gt}},\tau,I_0,\mathbf{a})
\end{equation}
to denoise the current chunk while attending only to its past. This bridges the architecture gap before few-step distillation. The resulting model is causal but still multi-step and is exposed to clean rather than self-generated history.

\paragraph{Stage 2: causal ODE initialization.}
A bidirectional teacher can access future frames that are unavailable to a causal student, creating a mismatch in their conditioning information.
Following Causal Forcing, we use the autoregressive diffusion model as the teacher for few-step initialization.
For guided distillation, we collect probability-flow ODE trajectories using CFG-guided predictions from this autoregressive teacher, and train a few-step generator $G_\phi$ to predict the corresponding clean endpoints from intermediate states at selected timesteps $\tau \in \mathcal{S}$:
\begin{equation}
\min_{\phi}\;\mathbb{E}\left[\left\|G_\phi\!\left(\mathbf{z}^{i}_{\tau},\mathbf{z}^{<i}_{\mathrm{gt}},\tau, I_0, \mathbf{a}\right)-\mathbf{z}^{i}_{0}\right\|_2^2\right].
\label{eq:causal_ode}
\end{equation}
Here, $\mathbf{z}^{i}_{0}$ denotes the teacher-generated endpoint for chunk $i$, and $\mathbf{z}^{<i}_{\mathrm{gt}}$ provides the ground-truth history.
Action conditioning is restricted to the current and preceding action windows.
This trajectory regression transfers the teacher's guided flow map into the single-branch student, jointly distilling the effect of CFG and the multi-step sampling process into a causal few-step initialization for subsequent distribution matching distillation.

\paragraph{Stage 3: asymmetric distribution matching.}
The causal initializer inherits the quality ceiling and exposure bias of the autoregressive teacher. We therefore self-roll out the few-step student to obtain $\widetilde{\mathbf{z}}$, perturb it to $\widetilde{\mathbf{z}}_\tau$, and apply asymmetric distribution matching distillation (DMD)~\cite{yin2024dmd,yin2025bidirectional}:
\begin{equation}
  \nabla_\phi\mathcal{L}_{\mathrm{DMD}}
  =-\mathbb{E}_{\widetilde{\mathbf{z}},\tau}
  \left[
  \left(s_{\mathrm{real}}(\widetilde{\mathbf{z}}_\tau,\tau,I_0,\mathbf{a})-
  s_{\mathrm{fake}}(\widetilde{\mathbf{z}}_\tau,\tau,I_0,\mathbf{a})\right)
  \frac{\partial\widetilde{\mathbf{z}}}{\partial\phi}
  \right].
  \label{eq:asymmetric_dmd}
\end{equation}
Here $s_{\mathrm{real}}$ denotes the CFG-guided score supplied by the frozen high-quality bidirectional teacher, while $s_{\mathrm{fake}}$ denotes the conditional score of the student rollout distribution estimated by an online diffusion model.
The asymmetry is intentional: generation remains causal, whereas the real-score teacher may use bidirectional context to pull the complete rollout toward the higher-quality target distribution. Training on self-generated histories exposes the student to its deployment distribution and directly addresses autoregressive error accumulation. The resulting model combines a causal action interface, chunk-wise rollout, and few denoising evaluations per chunk for low-latency interaction.

\section{Physical Data Pipeline}
\label{sec:data_pipeline}

Physical world models learn from visual observations of objects moving, interacting, and changing state. Network videos capture diverse physical processes, but editing transitions can interrupt an interaction, persistent overlays can obscure it, and poor image quality can hide the relevant visual evidence. Preparing these videos for training therefore requires both clip curation and descriptions that identify the objects, actions, and temporal progression of each physical event.

\begin{figure}[!tp]
\centering
\includegraphics[width=\linewidth]{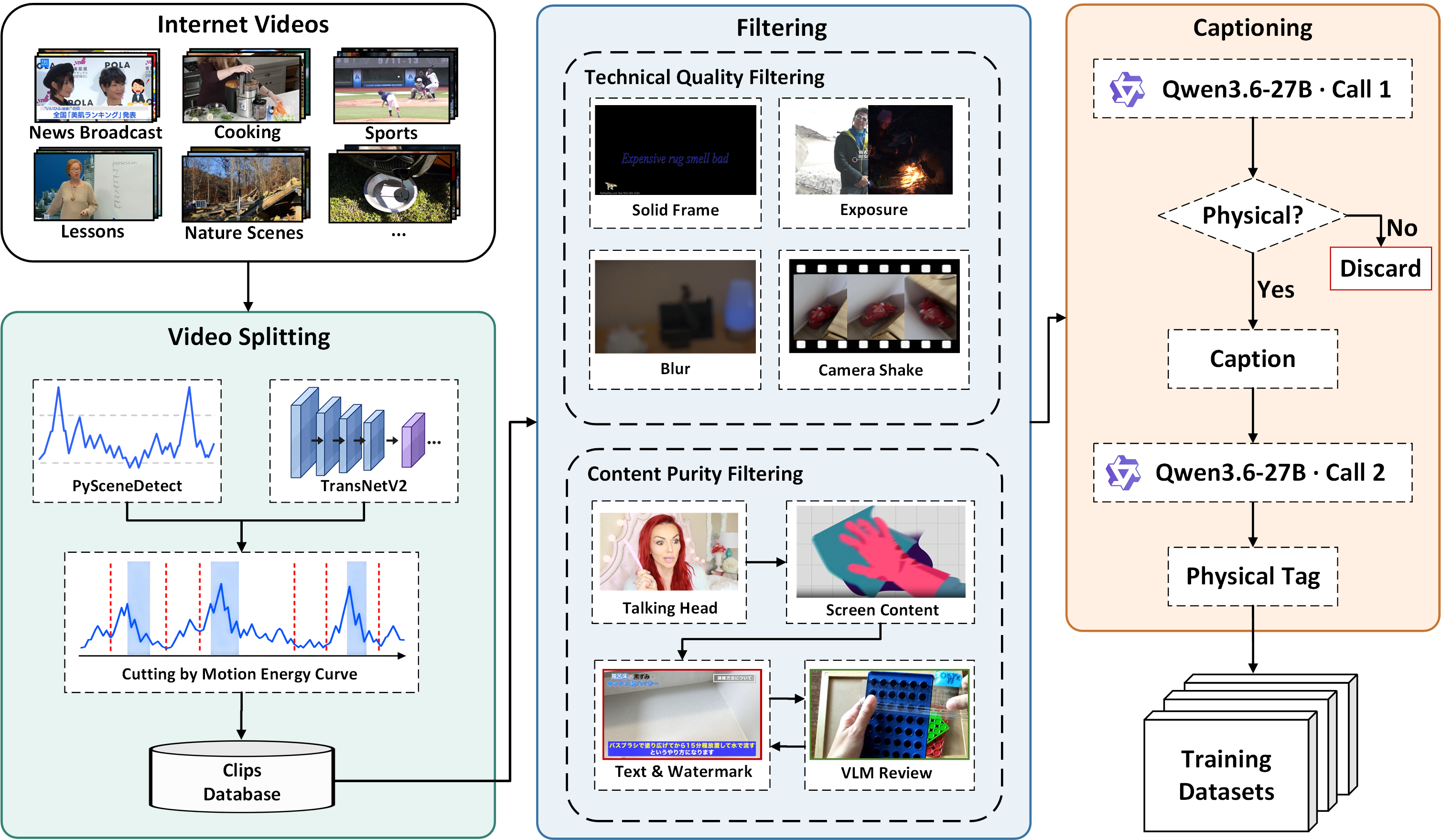}
\caption{Physical data pipeline. Source videos undergo Spatiotemporal Video Splitting, Technical Quality Filtering, and Content Purity Filtering. Physical Video Captioning first performs physical verification and generates a structured caption, then produces physical tags in a second model call. The resulting clips and annotations provide candidates for final dataset selection.}
\label{fig:architecture}
\end{figure}

Our pipeline combines clip curation with structured captioning and tagging (Figure~\ref{fig:architecture}). Section~\ref{sec:data_sourcing} describes the source collection, and Section~\ref{sec:video_curation} explains how source videos are split and filtered into candidate clips. Section~\ref{sec:physical_caption} presents physical verification and the two annotation calls, while Section~\ref{sec:physical_taxonomy} defines the accompanying tagging record. Section~\ref{sec:data_statistics} summarizes the annotated outputs and their use in final dataset selection.

\subsection{Data Sourcing}
\label{sec:data_sourcing}

We assemble the input collection from two sources: agentic retrieval and domain-specific providers. For agentic retrieval, a sourcing agent uses the physical taxonomy (Section~\ref{sec:physical_taxonomy}) to expand queries across physical interactions, environments, and manipulation tasks. Queries cover processes such as granular flows, multi-body impacts, tool use, and elastic deformation. An initial relevance filter screens the retrieved videos before clip curation.

Domain-specific providers contribute physical-interaction and robotics footage that complements the retrieved collection. Together, these sources provide diverse input video for the curation pipeline.

\subsection{Video Curation Pipeline}
\label{sec:video_curation}

The Video Curation Pipeline begins with Spatiotemporal Video Splitting (Section~\ref{sec:splitting}) to produce candidate clips. Technical Quality Filtering (Section~\ref{sec:tech_quality}) and Content Purity Filtering (Section~\ref{sec:content_purity}) then remove visual defects and distracting content before physical verification and annotation.

\subsubsection{Spatiotemporal Video Splitting}
\label{sec:splitting}

Uniform splitting can cut through an interaction or combine frames from different shots. We use Shot Boundary Detection to identify continuous shots, followed by Motion-Aware Fixed Window Selection to extract fixed-duration clips within each shot while favoring boundaries with low motion.

\paragraph{Shot Boundary Detection.}
We combine PySceneDetect~\cite{pyscenedetect} with TransNetV2~\cite{soucek2020transnetv2} to detect abrupt cuts and gradual transitions. Photometric changes provide candidate boundaries, while TransNetV2 captures transitions across multiple frames. Nearby detections are consolidated into transition intervals. Comparing frames before and after each interval suppresses transient flashes that return to the same scene. A short guard interval on either side of a confirmed transition excludes blended frames. Subsequent clip selection stays within the resulting shot boundaries.

\paragraph{Motion-Aware Fixed Window Selection.}
Within each shot, we compute motion energy as the temporally smoothed mean magnitude of Farneb\"ack optical flow~\cite{farneback2003flow}. The flow computation restarts at each shot boundary. This signal guides the placement of fixed-duration clips: lower motion at the endpoints reduces action truncation, while stronger motion within a clip favors active physical processes.

For a target duration $\ell$, we divide a shot into non-overlapping candidate intervals, each large enough to contain one clip. Within a candidate interval $[a,b]$, the clip start time is selected by
\begin{equation}
u^{*} = \underset{a \leq u \leq b-\ell}{\operatorname{arg\,min}}
\left[B(u) - \lambda_c A(u) + \lambda_p D(u)\right].
\label{eq:fixed_window_selection}
\end{equation}
Here, $B(u)$ is the normalized motion energy at the two clip endpoints, $A(u)$ is the normalized motion energy accumulated within the clip, and $D(u)$ is the squared normalized distance from the center of the feasible start-time range. The weights $\lambda_c$ and $\lambda_p$ balance motion coverage and placement. Each selected clip remains inside its candidate interval, so clips do not overlap or cross shot boundaries. Shots shorter than the target duration are discarded, and residual footage outside complete windows is omitted.

Fixed-duration selection is the primary mode used in this pipeline. As an alternative, the splitter supports a variable-duration mode in which dynamic programming partitions each shot under minimum and maximum duration constraints, balancing clip lengths while penalizing cuts through active motion.

\subsubsection{Technical Quality Filtering}
\label{sec:tech_quality}

\begin{figure}[!tp]
\centering
\includegraphics[width=0.95\linewidth]{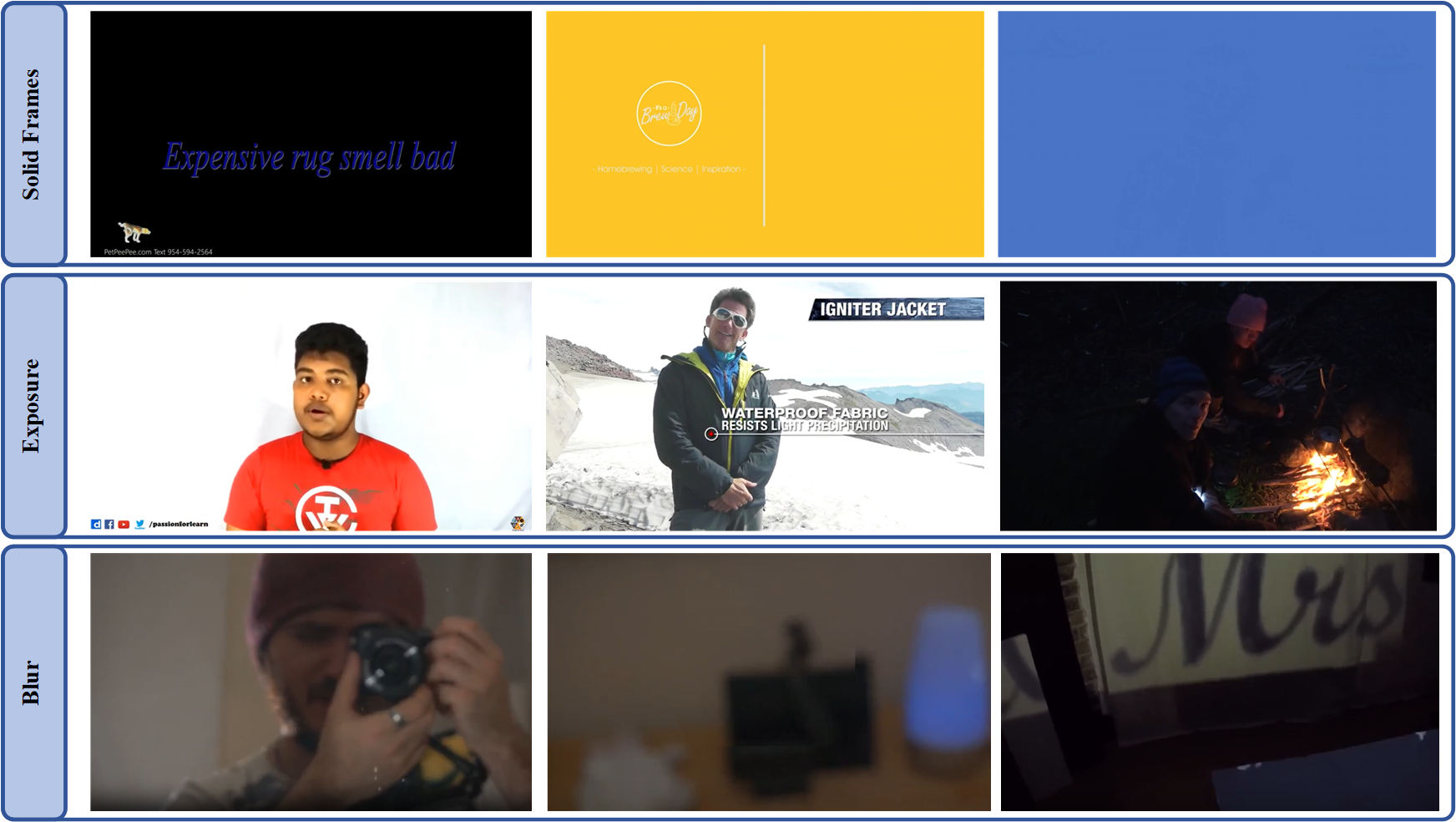}
\caption{Examples of visual defects addressed by Technical Quality Filtering: solid-color frames, severe exposure defects, and persistent blur. Input-validity and temporal-quality checks are described in the text.}
\label{fig:technical-quality}
\end{figure}

Clips with valid temporal boundaries may still contain visual defects that obscure physical interactions. Technical Quality Filtering checks Input Validity, Visual Quality, and Temporal Quality. Representative visual defects addressed by these checks are shown in Figure~\ref{fig:technical-quality}.

\textbf{Input Validity.} We first verify that the clip can be decoded and that its resolution and duration meet the input requirements. Clips failing these checks are rejected before further analysis.

\textbf{Visual Quality.} For black frames, solid-color frames, and exposure defects, we measure grayscale intensity, spatial pixel variation, and the proportions of dark and saturated pixels across sampled frames. Their prevalence over time distinguishes clips dominated by uninformative or poorly exposed frames from clips containing brief lighting changes. Persistent defects cause rejection. Blur is assessed using the distribution of Laplacian variance across frames: a clip is rejected when both its median sharpness and lower-tail sharpness are low, preserving fast interactions that contain only brief motion blur.

\textbf{Temporal Quality.} We detect frame stuttering and camera shake. Consecutive-frame differences identify near-duplicate frames, and clips with a high proportion of repeated frames are rejected. For camera shake, Lucas--Kanade feature tracking~\cite{lucas1981registration} and RANSAC~\cite{fischler1981ransac} estimate interframe camera motion. Subtracting a smoothed camera trajectory isolates rapid translation, rotation, and scale fluctuations. Clips with strong, persistent residual jitter are rejected, while smooth pans and tracking shots are retained.

Checks are applied from inexpensive validation to more costly motion analysis, stopping once a clip meets a rejection condition. Retained clips proceed to Content Purity Filtering.

\subsubsection{Content Purity Filtering}
\label{sec:content_purity}

Clear, stable clips can still be dominated by talking heads, screen recordings, or post-production overlays. Content Purity Filtering uses four components: Face and Talking-Head Filtering, Screen Content Screening, Text and Watermark Detection, and VLM Text Review. The first three identify unwanted content or trigger further review; VLM Text Review resolves ambiguous text and screen-content detections.

\begin{figure}[!t]
\centering
\includegraphics[width=0.96\linewidth]{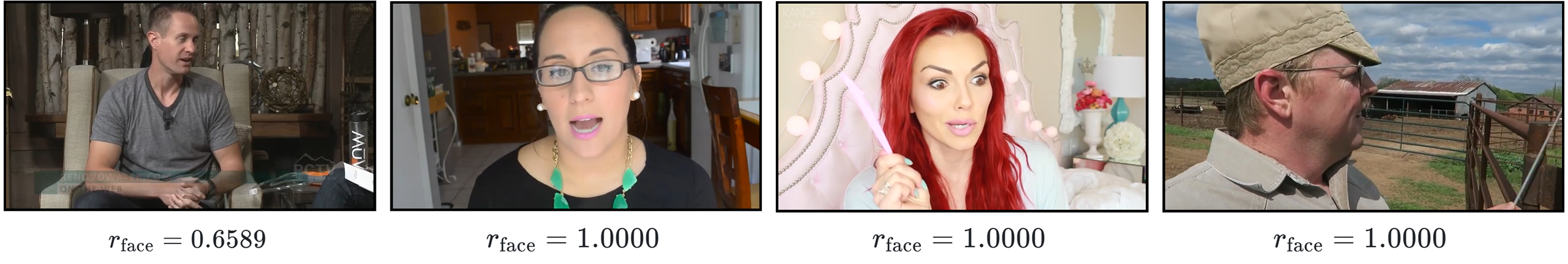}
\caption{Face and Talking-Head Filtering rejects presenter-dominated clips while retaining clips with distant operators and human--object interactions.}
\label{fig:face-filter}
\end{figure}

\paragraph{Face and Talking-Head Filtering.}
YuNet~\cite{wu2023yunet} detects faces and facial landmarks across sampled frames. Face orientation, area, and temporal persistence identify clips dominated by frontal presentations and interviews. Combined with talking-head cues, these measurements distinguish presenter-focused footage from clips in which an operator participates in an object interaction (Figure~\ref{fig:face-filter}).

\paragraph{Screen Content Screening.}
A CLIP encoder~\cite{radford2021clip} compares sampled frames with prompts for screen recordings, presentation slides, and studio content, using physical-interaction prompts as a reference. Relative semantic scores identify screen-dominated content and route ambiguous UI detections to VLM Text Review.

\paragraph{Text and Watermark Detection.}
An EasyOCR-based text-detection branch~\cite{easyocr} detects text regions and tracks their position, size, and persistence across frames. Stable text tracks in frame corners identify screen-locked watermarks, while recurring text near frame borders identifies subtitle and banner candidates. Clear watermark detections cause rejection; other text detections proceed to VLM Text Review to distinguish overlays from text on physical objects.

\begin{figure}[!t]
\centering
\includegraphics[width=0.96\linewidth]{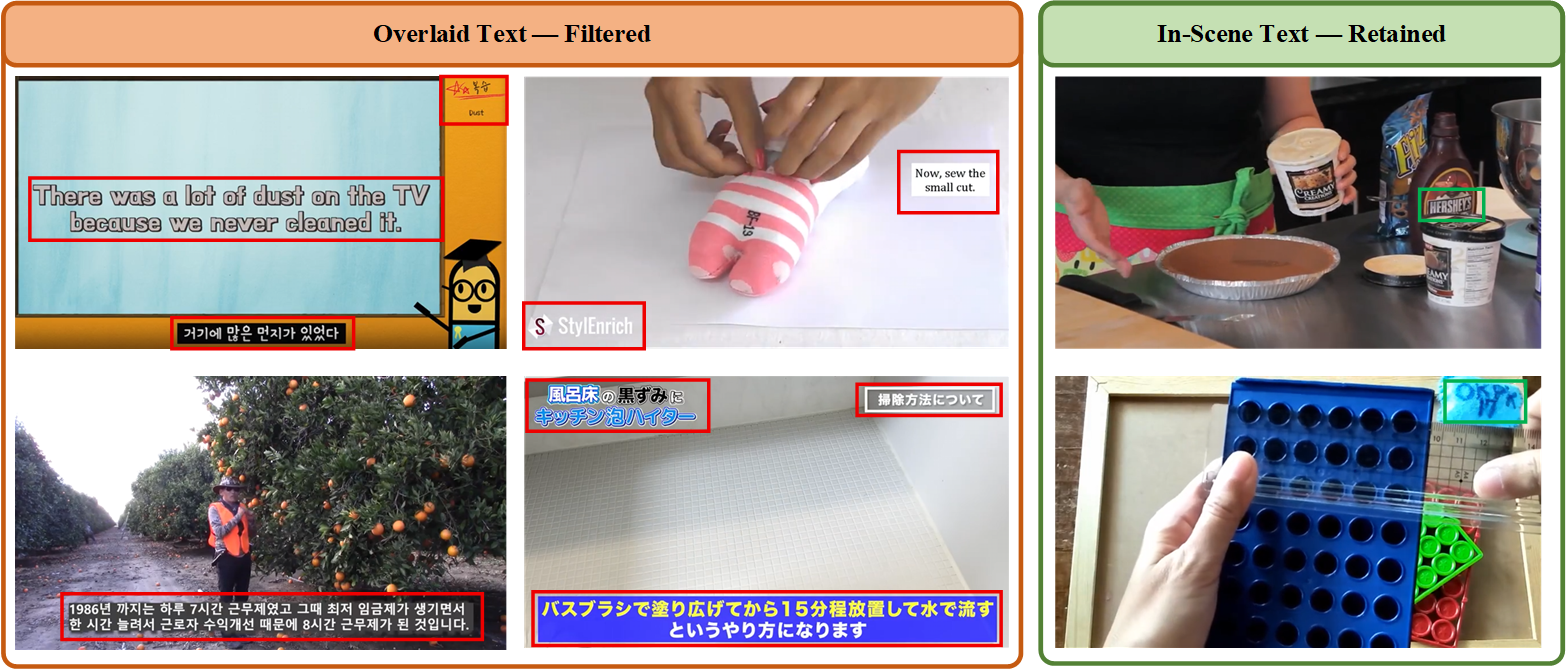}
\caption{VLM Text Review distinguishes text on physical objects from post-production overlays. Clips containing authentic scene text are retained for physical verification.}
\label{fig:text-disambiguation}
\end{figure}

\paragraph{VLM Text Review.}
Qwen3-VL-4B~\cite{bai2025qwen3vl} reviews six temporal keyframes arranged in a contact sheet and returns one of five classes: \texttt{scene\_text}, \texttt{overlay\_ui}, \texttt{both}, \texttt{no\_text}, or \texttt{uncertain}. Clips classified as \texttt{scene\_text} or \texttt{no\_text} continue to physical verification; the remaining classes are rejected. This step preserves labels, gauges, and signs that belong to the physical scene while removing post-production text and UI (Figure~\ref{fig:text-disambiguation}).

\subsection{Physical Video Captioning}
\label{sec:physical_caption}

\begin{figure}[ht]
\centering
\includegraphics[width=0.95\linewidth]{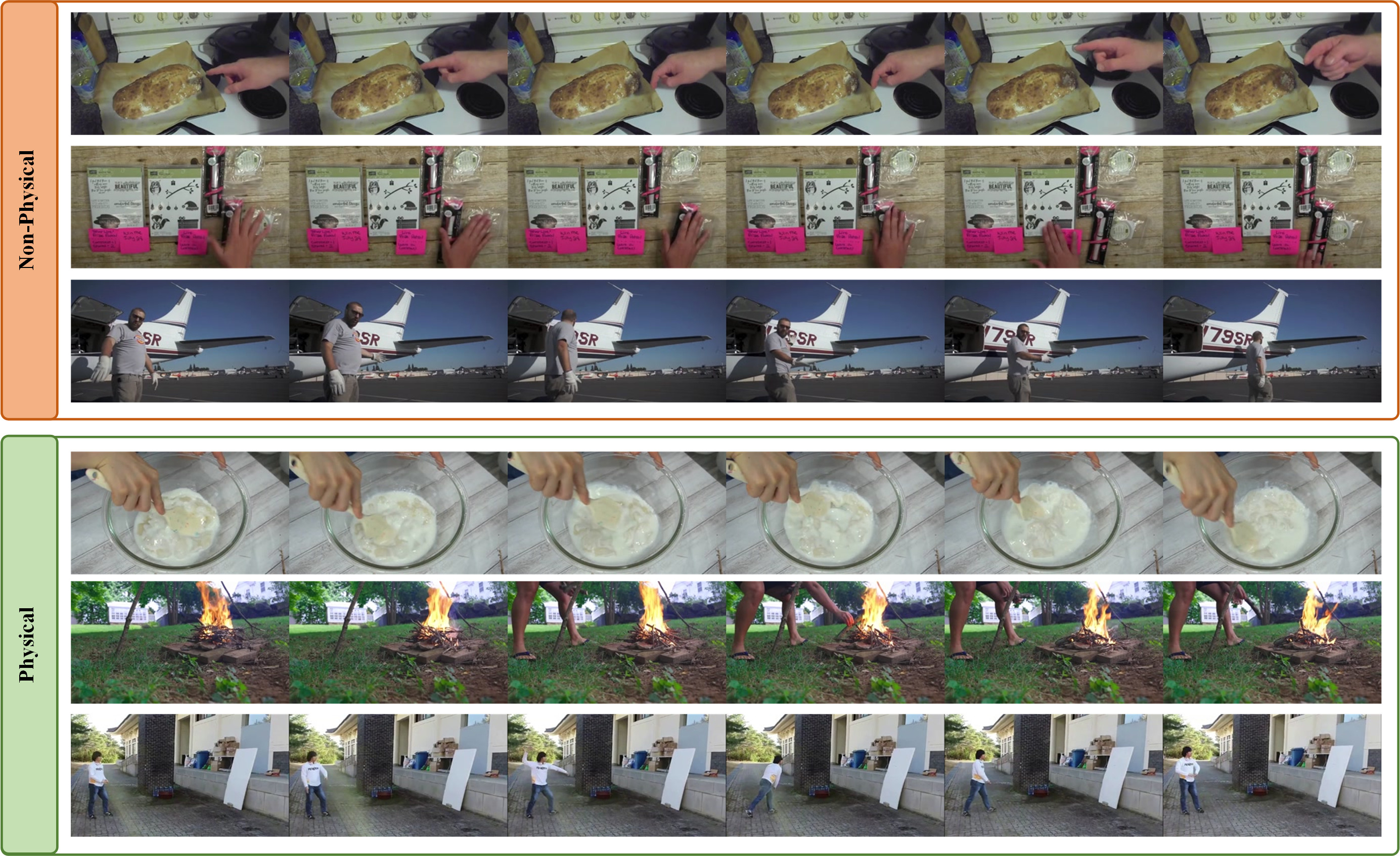}
\caption{Physical verification in the first annotation call. Top row: rejected clips dominated by static product staging and decorative hand gestures. Bottom row: retained clips containing physical interactions. Retained clips receive complete structured captions and proceed to Closed-Set Tag Generation.}
\label{fig:physical_verification}
\end{figure}

Content Purity Filtering removes distracting content, while Physical Video Captioning determines whether a clip contains physical processes and describes their progression. Qwen3.6-27B~\cite{qwen2026qwen36_27b} produces a structured caption with four information groups:
\begin{enumerate}
\item \textbf{Global Scene Dynamics}: The environment, initial scene, and progression of the main physical interaction.
\item \textbf{Camera Motion and Illumination}: Camera movement, shot scale, viewing angle, composition, and lighting, described separately from object motion.
\item \textbf{World Knowledge and Material Properties}: Relevant scene knowledge, object materials, and spatial relationships grounded in the clip.
\item \textbf{Entity-Level Temporal Behaviors}: The participating objects and actors, with timestamped actions describing contact, manipulation, deformation, motion, and changes of state.
\end{enumerate}

Appendix~\ref{app:structured-caption-example} presents a real structured-caption example using field names aligned with these four information groups. It also shows the exact compact JSON object serialized as the input to the StrucPhysVideo text encoder.

To produce these descriptions and the accompanying tags, Frame Preparation establishes the shared inputs, Two-Call Annotation generates the captions and tags, and Output Validation checks the resulting records.

\paragraph{Frame Preparation.}
We uniformly extract 24 frames across each clip and retain their native timestamps. Frames are resized to a maximum long-edge dimension of 768 pixels. The timestamped frames and clip metadata are stored in an immutable cache and used by both annotation calls.

\paragraph{Two-Call Annotation.}
The first call verifies physical relevance and generates the caption; clips that pass this check proceed to the second call for tag generation:
\begin{enumerate}
\item \textbf{Physical Verification and Caption Generation.} Qwen3.6-27B receives the timestamped frames, clip metadata, and domain tags. It returns the Boolean flag \texttt{is\_physics\_related} with supporting visual evidence and, for a physical clip, the complete structured caption containing the four information groups above. Clips dominated by static product staging or decorative gestures are rejected at this step (Figure~\ref{fig:physical_verification}).
\item \textbf{Closed-Set Tag Generation.} For clips passing physical verification, a second call receives the cached timestamped frames and clip metadata, together with the first-call caption. It generates tags using the closed-set taxonomy and attribute vocabularies described in Section~\ref{sec:physical_taxonomy}. The frames remain the primary evidence when the caption or external metadata differs from the visible content. The tags are merged with the caption into one annotation record.
\end{enumerate}


\paragraph{Output Validation.}
Validation checks inference completion, JSON structure, required fields, closed-set tag values, and action timestamps within the clip duration. The annotation manifest aligns each retained clip with its caption and tags before final dataset selection.

\subsection{Physical Taxonomy and Structured Tagging}
\label{sec:physical_taxonomy}

For clips retained after video curation and physical verification, captions describe the scene, participating objects, and temporal progression of events. Organizing these clips into a training dataset also requires consistent categories for retrieval and composition analysis. Since free-form captions may describe the same activity or phenomenon in different terms, we complement them with structured tags drawn from predefined vocabularies.

The tags cover \textbf{general video attributes} and \textbf{physics-specific attributes}. General attributes describe content categories and photographic properties, while physics-specific attributes describe physical phenomena, materials, and state changes. Section~\ref{sec:physical_caption} presents the annotation procedure; this section defines the tags and their use in dataset organization.

\subsubsection{General Video Tags}

General video tags comprise one content-category field and seven photographic fields (Table~\ref{tab:general_video_tags}). The content category identifies the setting or activity, such as a laboratory demonstration, cooking, or industrial operation. Photographic fields record color, shot scale, viewing angle, lens type, composition, lighting characteristics, and light-source type.

\begin{table}[ht]
\centering
\footnotesize
\setlength{\tabcolsep}{4pt}
\renewcommand{\arraystretch}{1.12}
\caption{General video attributes, vocabulary sizes, and representative values.}
\label{tab:general_video_tags}
\begin{tabular}{@{}p{0.22\linewidth}cp{0.54\linewidth}@{}}
\hline
\textbf{Attribute} & \textbf{Vocabulary Size} & \textbf{Representative Values} \\
\hline
Content category
& 14
& Laboratory demonstration, educational explanation, daily life, industrial machinery, natural phenomena \\

Color
& 14
& Warm, cool, saturated, neutral, high-contrast, monochrome \\

Shot scale
& 7
& Extreme close-up, close-up, medium shot, full shot, long shot \\

Viewing angle
& 6
& Eye-level, low-angle, high-angle, overhead, Dutch angle \\

Lens type
& 5
& Ultra-wide, wide-angle, standard, telephoto, macro \\

Composition
& 6
& Rule of thirds, centered, symmetrical, diagonal, leading lines \\

Lighting characteristics
& 10
& Natural sunlight, soft diffuse light, harsh direct light, low-key lighting, backlighting \\

Light-source type
& 10
& Natural, incandescent, fluorescent, LED point source, studio strobe, firelight \\
\hline
\end{tabular}
\end{table}

These tags describe the video's content and presentation, but do not identify its physical processes. For example, the category \texttt{industrial-machinery} does not distinguish rotation from cutting or deformation. Physics-specific tags provide these distinctions.

\FloatBarrier

\subsubsection{Physics-Specific Attributes}

\begin{figure}[ht]
\centering
\includegraphics[width=\linewidth]{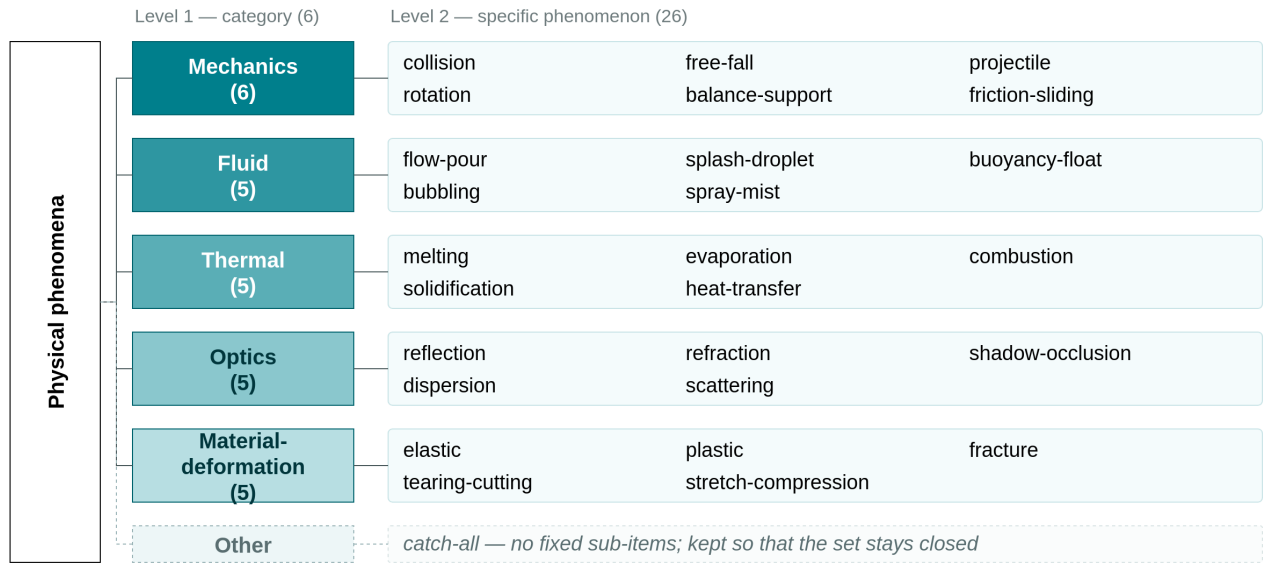}
\caption{Two-level taxonomy of physical phenomena.
Five top-level categories contain 26 predefined leaf phenomena.
An additional \emph{Other} category covers physical interactions
outside the vocabulary and has no predefined leaves.}
\label{fig:physical_taxonomy}
\end{figure}

\paragraph{Physical phenomenon taxonomy.}
The taxonomy has two levels: top-level categories and leaf phenomena.
Five top-level categories---Mechanics, Fluid Dynamics, Thermal Dynamics,
Wave \& Optics, and Material Deformation---contain 26 predefined leaf
phenomena. An additional category, \emph{Other}, covers physical
interactions outside this vocabulary and has no predefined leaves.
Figure~\ref{fig:physical_taxonomy} shows the category-to-leaf mapping.

\FloatBarrier

\paragraph{Multi-label annotation.}
A clip may receive multiple labels at both taxonomy levels because it
can contain several processes, either simultaneously or at different
stages. A primary phenomenon is designated when a single category
assignment is needed for statistics or sampling; the remaining labels
are retained. The tags record which phenomena occur, while the
timestamped caption describes their temporal progression and the
participating objects.

\begin{figure}[!tp]
\centering
\includegraphics[width=\linewidth]{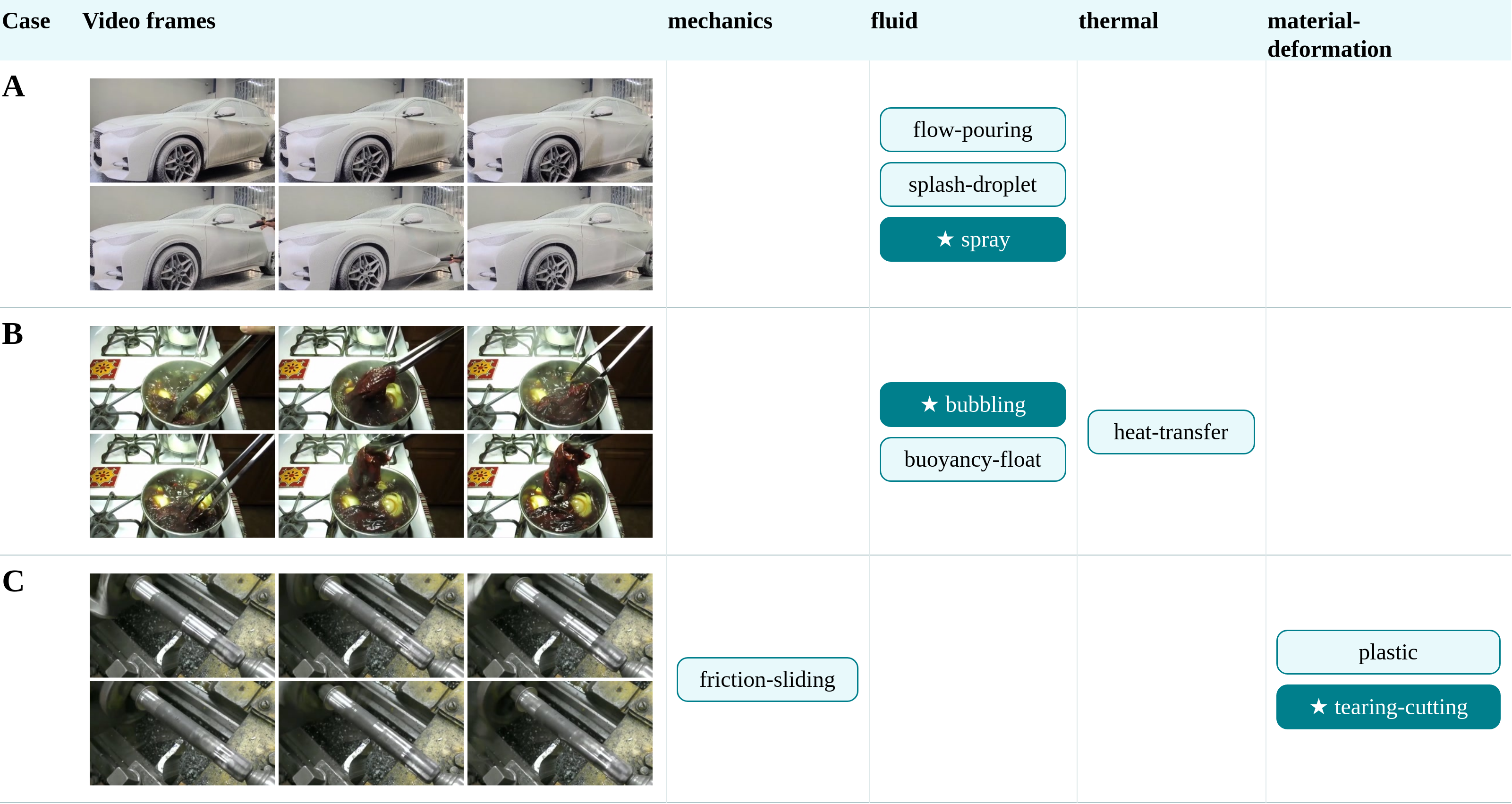}
\caption{Multi-label physical annotations for car washing, boiling
water, and lathe cutting. Each example contains six frames and the
associated phenomenon tags, grouped by top-level category.
A filled star marks the primary phenomenon; outlined tags indicate
additional phenomena.}
\label{fig:physical_tag_composition}
\end{figure}

Figure~\ref{fig:physical_tag_composition} shows annotations for car washing, boiling water, and lathe cutting.
In the lathe-cutting example, workpiece rotation and the changes produced by cutting occur within the same clip.
A content label identifies the activity as an industrial operation, whereas the phenomenon labels distinguish the processes within it.
The primary label provides a single attribution without replacing this multi-label description.

\paragraph{Additional physical attributes.}
Material categories include rigid solids, deformable solids, liquids,
and granular materials. State-change labels describe changes such as
deformation, phase transition, and fragmentation. These attributes
complement phenomenon labels by describing the participating materials
and the resulting changes. Interaction viewpoint distinguishes clips
without human participation from first- and third-person interaction
views.

Physical anomaly flags record visible behaviors such as object
interpenetration, unrealistic collisions, and gravity-defying motion.
They differ from the physical relevance indicator returned by the
first annotation call: a clip may contain a physical interaction while
depicting that interaction implausibly. The judgment concerns visible
behavior rather than whether the video is rendered or generated.

We exclude absolute mass, force magnitude, real-world speed, and temperature from the annotation.
For videos without the necessary scale, calibration, or additional measurements, the annotator cannot reliably determine these quantities.
Requesting them would encourage unsupported estimates and introduce annotation noise.
The tags instead describe physical phenomena, material categories, and visible state changes.
Image-based motion measurements are retained separately and are not interpreted as estimates of physical energy.

The content and phenomenon tags allow dataset composition to be examined by both scene type and physical process.
For subsequent selection, they remain associated with the quality and motion measurements retained from video curation.
Section~\ref{sec:data_statistics} summarizes the annotated candidates and their use in selecting data to match the distribution of a reference pretraining corpus.

\FloatBarrier

\subsection{Curation Outcome}
\label{sec:data_statistics}

\begin{figure}[!tp]
\centering
\includegraphics[width=\linewidth]{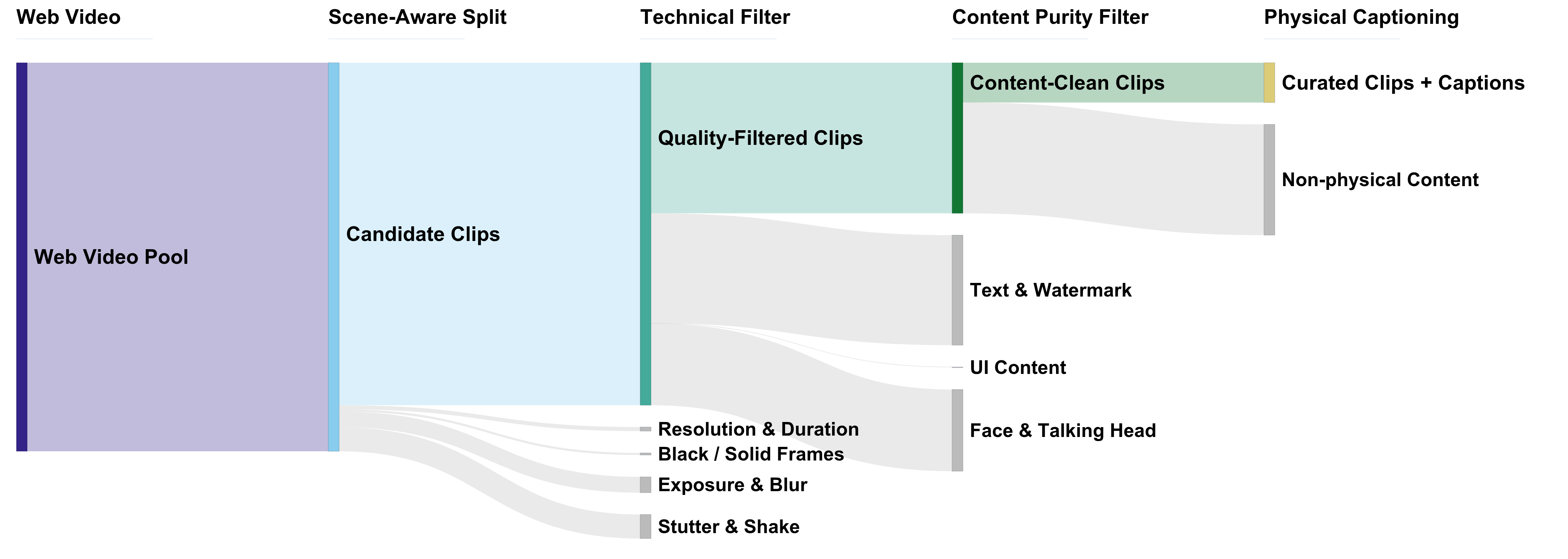}
\caption{Curation from source videos to annotated clips. Spatiotemporal Video Splitting produces candidate clips; Technical Quality Filtering and Content Purity Filtering select clips for Physical Video Captioning. Physical Verification selects physical clips, which receive structured captions and closed-set tags.}
\label{fig:data_curriculum}
\end{figure}

Figure~\ref{fig:data_curriculum} summarizes clip filtering at three checkpoints: Technical Quality Filtering, Content Purity Filtering, and Physical Verification. Technical Quality Filtering retains clips with usable visual and temporal quality. Content Purity Filtering removes presenter-dominated footage and overlays while preserving text that belongs to the scene. Physical Verification then selects clips containing physical processes for captioning and tagging.

The annotation-stage output pairs each retained clip with a structured caption and a tagging record. Captions describe the scene, objects, and temporal progression of interactions; tagging records organize their categorical attributes, curation measurements, and derived information.

These annotated candidates are used in a final dataset-selection step for distribution matching against a reference pretraining corpus. Tags, sampling weights, and deduplication results support this selection, linking the clip-level records to the composition of the resulting training data.

\subsection{Action-Conditioned Data Curation}
\label{sec:action_data_curation}

\begin{figure}[!tp]
\centering
\includegraphics[width=\linewidth]{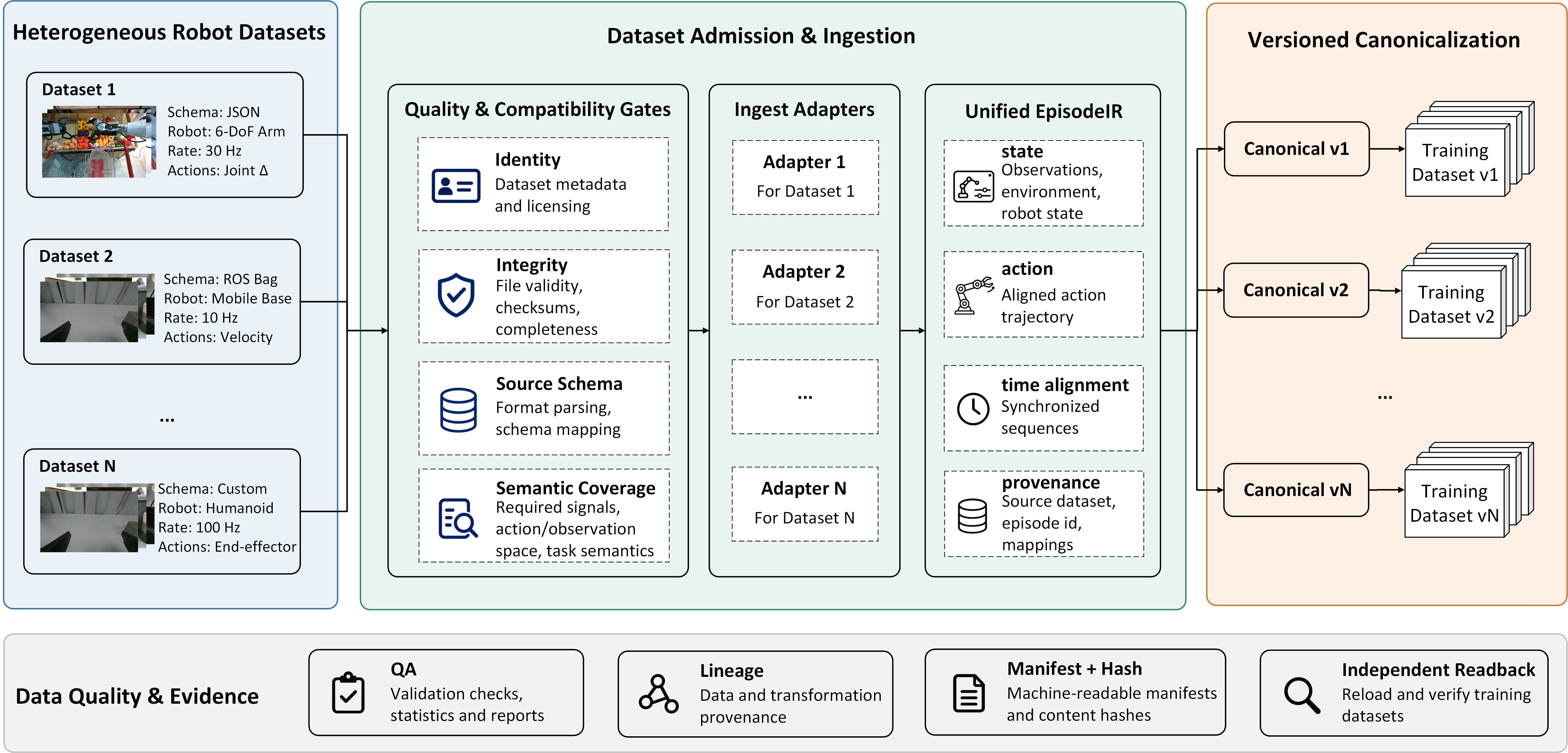}
\caption{Action-conditioned data curation. Dataset-specific gates and adapters produce a unified EpisodeIR. Canonical versions define the training representations, with semantic checks during production and quality checks and independent readback before delivery of the corresponding training datasets.}
\label{fig:action_data_curation}
\end{figure}

To support action-conditioned training across heterogeneous robot datasets,
we organize data preparation into Data Ingestion, Canonical Production,
and Output Validation (Figure~\ref{fig:action_data_curation}).

Data Ingestion applies dataset-specific admission checks to dataset identity, release metadata, input integrity, source schema, and semantic coverage. Dedicated ingest adapters discover episodes, map source fields, align temporal streams, interpret actions and observations, and preserve provenance. The resulting EpisodeIR provides a common representation of each episode and a stable semantic boundary between source formats and downstream training representations.

Canonical Production converts EpisodeIR into versioned training datasets. Each Canonical version specifies its field layout, action semantics, masks, temporal sampling and terminal-frame policies, and normalization strategy. Semantic checks verify gripper values, rotations, and action targets, together with their alignment to observations.

Output Validation checks numerical validity, field and mask consistency, normalization statistics, and accounting of retained and excluded episodes. Each release includes a manifest, lineage records, and quality-check results. Independent readback verifies persisted data and compatibility with the training loader. Exclusions and processing failures are recorded against their source episodes, while accepted data provide aligned robot observations and action trajectories for training.

\section{Experiments}
\label{sec:experiments}

We evaluate the model in two complementary settings. Text-image-to-video (TI2V) evaluation measures the physical plausibility of generated dynamics under image and language conditioning, whereas image-action-to-video (IA2V) evaluation tests whether robot rollouts follow action trajectories while preserving visual fidelity.

\subsection{Text-Image-to-Video Evaluation}
\label{sec:ti2v_experiments}

\subsubsection{Implementation Details}
\label{sec:ti2v_implementation}
Starting from the official LingBot-Video checkpoint~\cite{ma2026lingbotvideo}, we trained the StrucPhysVideo-TI2V 30B backbone on 81-frame clips at 15 fps and $480\times832$ resolution. Training used four eight-H200 nodes with local and global batch sizes of 3 and 96. We combined eight-way expert parallelism, FSDP2 sharding of non-expert parameters, DeepEP token dispatch, non-reentrant activation checkpointing, and packed variable-length attention.

The corpus comprised 10,000 hours each of physics-focused videos and high-quality general-domain videos collected from the Internet. A dynamic schedule sampled the general-domain and physics-focused pools at 70\%/30\% during the first 60\% of training, 40\%/60\% during the next 30\%, and 20\%/80\% during the final 10\%. This progression preserved general-purpose generation while increasing the emphasis on physical dynamics.

Training retained FP32 master parameters and optimizer states, used BF16 forward computation, FP32 gradient reduction and sensitive operations, and TF32 matrix multiplication where supported. AdamW used a learning rate of $1\times10^{-5}$, weight decay of $1\times10^{-2}$, $\beta=(0.9,0.999)$, and a gradient-norm limit of 1.0. The cosine schedule used 100 warm-up steps and a $1\times10^{-6}$ minimum learning rate. Flow matching used 1,000 timesteps, a shift of 5, and logit-normal timestep sampling. The Qwen3-VL condition encoder and Wan VAE remained frozen in evaluation mode.

\subsubsection{Performance Comparison}
\label{sec:ti2v_performance}


\begin{figure}[!t]
  \centering
  \includegraphics[width=0.99\linewidth]{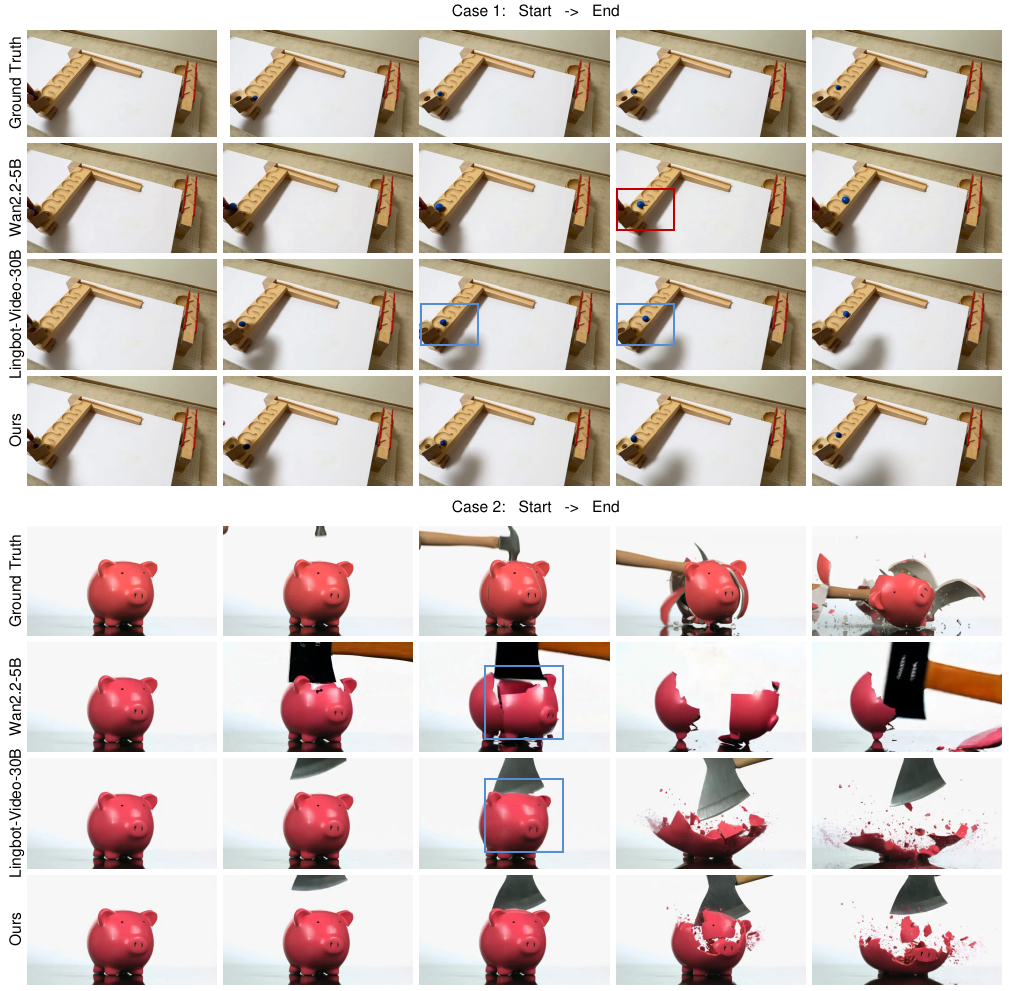}
  \caption{Qualitative comparison among ground truth, Wan2.2-5B\cite{wan2025}, LingBot-Video-30B\cite{ma2026lingbotvideo}, and StrucPhysVideo (Ours), with five frames shown from the start to the end of each sequence. In Case 1, the red boxes highlight substantial deformation of the track geometry, while the blue boxes indicate that the ball fails to roll along the shape of the track. In contrast, our model generates physically plausible ball-rolling motion that faithfully follows the track geometry. In Case 2, the blue boxes highlight the loss of the subject's key facial features, whereas our model effectively preserves the subject's identity and distinctive appearance.}
  \label{fig:ti2v-three-model-comparison}
\end{figure}

Physics-IQ Verified evaluates the reproduction of real physical dynamics. Physics-IQ contains 66 experiments across solid and fluid dynamics, thermodynamics, optics, and magnetism; three viewpoints and two repetitions yield 396 videos~\cite{motamed2026physicsiq}. For 198 instances, models receive the 3-second frame and description and predict the next 5 seconds, while paired repetitions quantify natural variation. The Verified protocol corrects ambiguous prompts and spurious reference motion and averages scores per sample~\cite{radsch2026physicsiqverified}. Its score equally combines spatial IoU, spatiotemporal IoU, weighted spatial IoU, and inverse normalized pixel MSE after normalization by paired-video variation. Higher values indicate closer agreement in event location, timing, magnitude, and appearance~\cite{radsch2026physicsiqverified}.

\begin{figure}[!tp]
  \centering
  \includegraphics[width=0.99\linewidth]{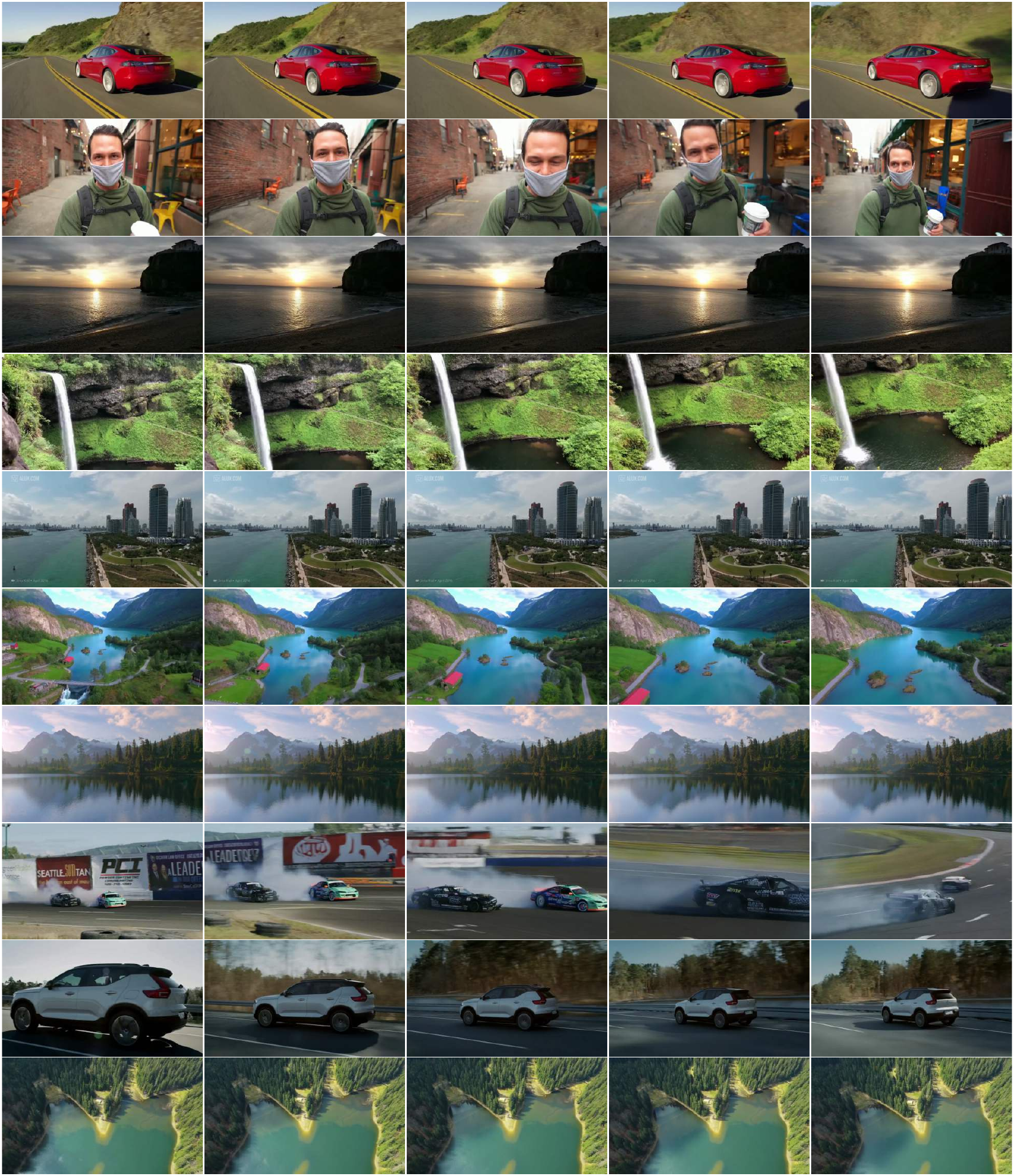}
  \caption{Qualitative StrucPhysVideo-TI2V results for general-domain video generation. Each row shows five temporally ordered frames from one generated video, spanning people, vehicles, landscapes, and camera motion.}
  \label{fig:ti2v-general-qualitative}
\end{figure}

\begin{figure}[!tp]
  \centering
  \includegraphics[width=0.99\linewidth]{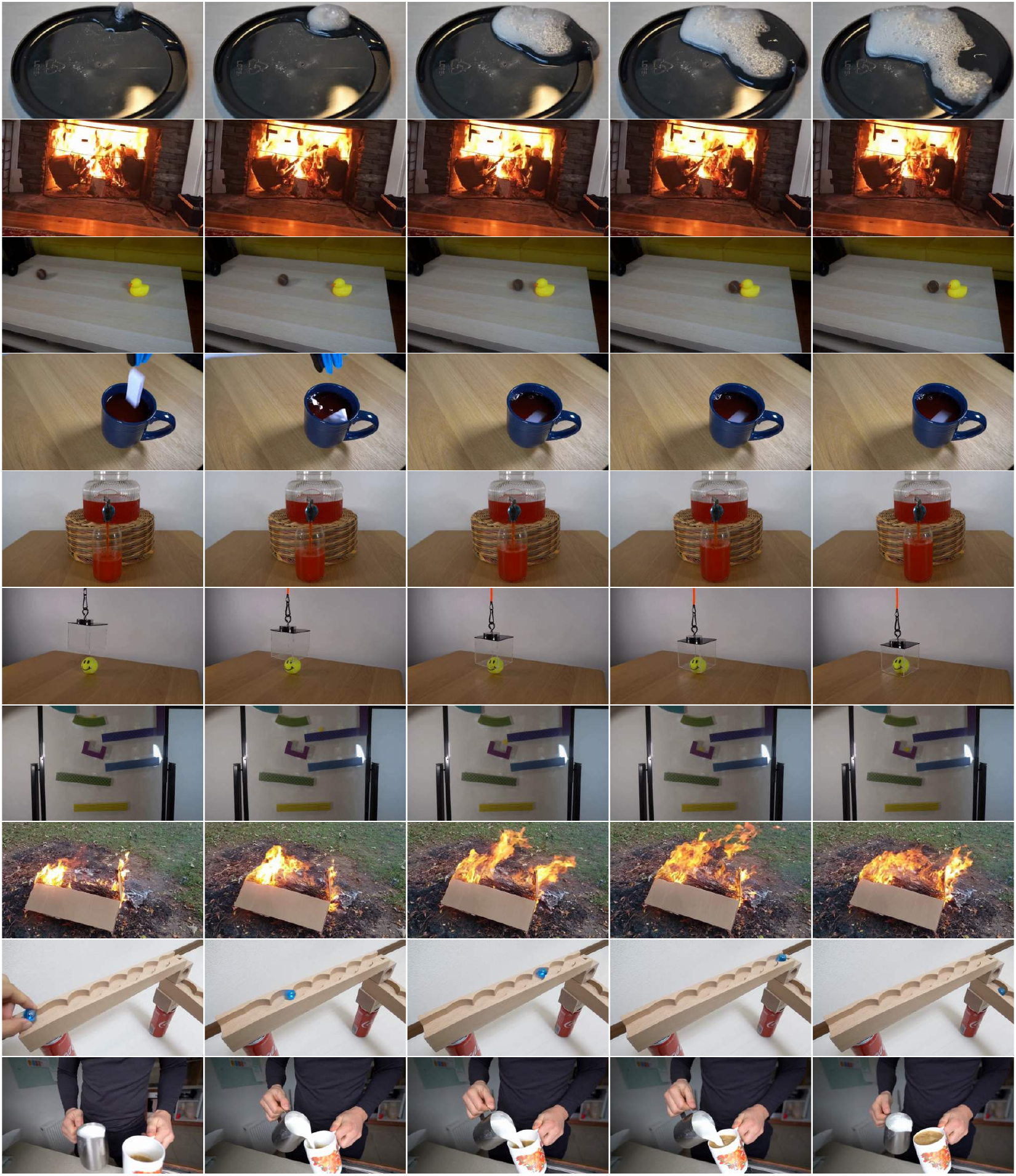}
  \caption{Qualitative StrucPhysVideo-TI2V results for physical-world video generation. Each row shows five temporally ordered frames from one generated video depicting a physical phenomenon or object interaction.}
  \label{fig:ti2v-physics-qualitative}
\end{figure}

\begin{figure}[!tp]
  \centering
  \includegraphics[width=0.99\linewidth]{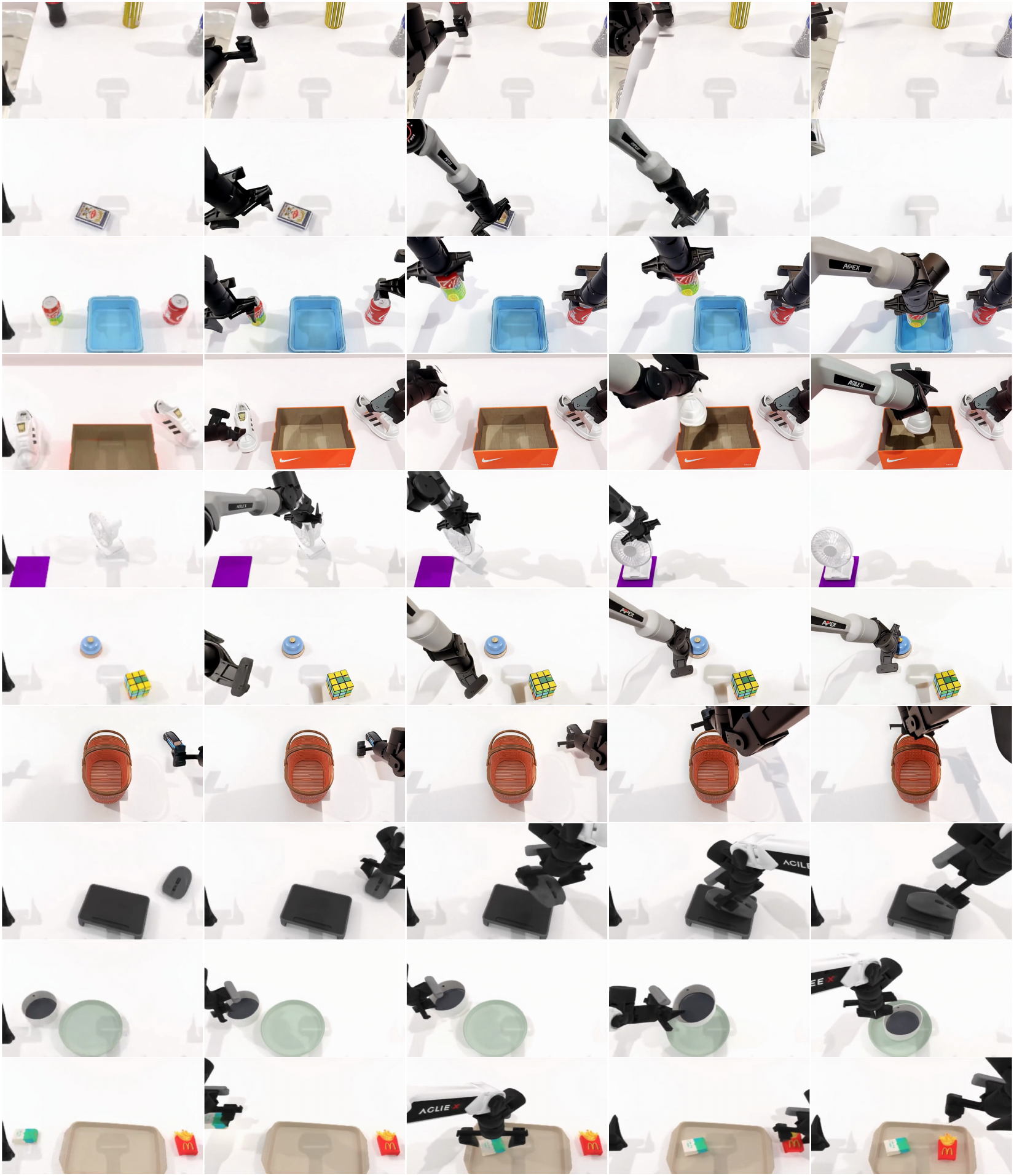}
  \caption{Qualitative StrucPhysVideo-TI2V results for embodied scenarios. Each row shows five temporally ordered frames from a generated robot manipulation sequence.}
  \label{fig:ti2v-embodied-qualitative}
\end{figure}

Under the verified image-to-video protocol, StrucPhysVideo achieved 45.5\%, the highest score among the 17 models in the comparison based on the 16 September 2026 benchmark snapshot (Figure~\ref{fig:physics_iq_ti2v}). We reproduced the LingBot-Video score, while all other comparison scores were taken directly from the benchmark snapshot. StrucPhysVideo was 2.8 percentage points above Cosmos3-Super Image2Video at 42.7\%~\cite{nvidia2026cosmos3} and 9.3 points above the strongest closed-source comparator, MiniMax H3 Max at 36.2\%~\cite{minimax2026h3,physicsiqleaderboard2026}. This result is specific to the evaluated protocol and leaderboard snapshot. Figure~\ref{fig:ti2v-three-model-comparison} compares StrucPhysVideo directly with Wan2.2-5B and LingBot-Video-30B. Additional examples span general-domain scenes (Figure~\ref{fig:ti2v-general-qualitative}), physical phenomena and interactions (Figure~\ref{fig:ti2v-physics-qualitative}), and embodied manipulation (Figure~\ref{fig:ti2v-embodied-qualitative}); these selected outputs illustrate breadth but do not replace quantitative evaluation.

\clearpage

\subsubsection{Ablation Studies}
\label{sec:ti2v_ablations}

\begin{figure}[!t]
  \centering
  \includegraphics[width=0.60\linewidth]{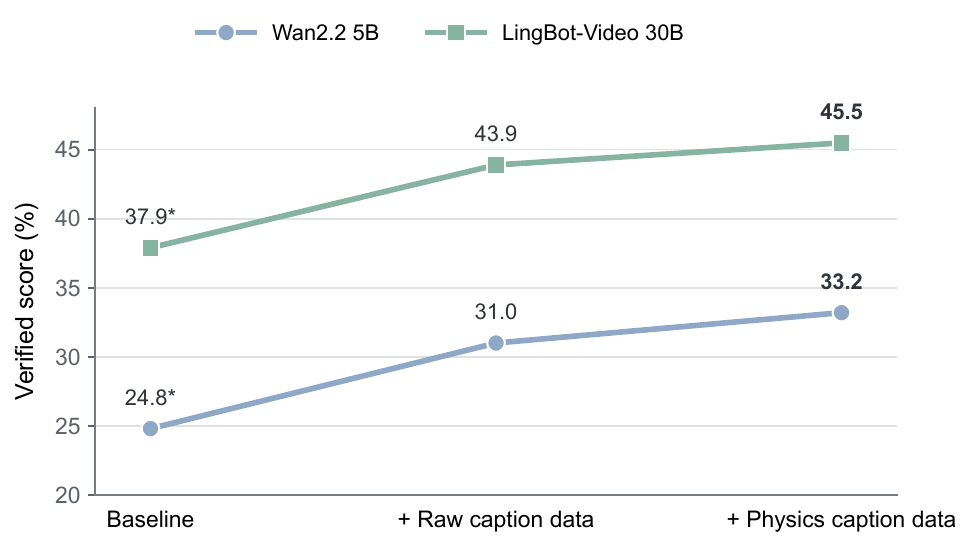}
  \caption{Caption ablation on Physics-IQ Verified. Wan2.2-5B~\cite{wan2025} and LingBot-Video-30B~\cite{ma2026lingbotvideo} are evaluated under three settings: the reproduced baseline, incorporation of raw captions, and incorporation of physics-focused captions. Physics-focused captions yield the highest Verified Score for both backbones. Only Physics-IQ Verified scores are reported. $^{*}$ denote scores reproduced by us.}
  \label{fig:physics_caption_ablation}
\end{figure}

To isolate the effect of caption supervision, we conducted an ablation on Wan2.2-5B~\cite{wan2025} and LingBot-Video-30B~\cite{ma2026lingbotvideo} using data prepared with our pipeline (Figure~\ref{fig:physics_caption_ablation}). For the Case~1 video in Figure~\ref{fig:ti2v-three-model-comparison}, Appendix~\ref{app:structured-caption-example} (Listing~\ref{lst:structured-caption-example}) provides the real structured caption, and Appendix~\ref{app:raw-caption-example} (Listing~\ref{lst:raw-caption-example}) provides the corresponding real raw caption. Adding raw captions increased the Physics-IQ Verified Score from 24.8\% to 31.0\% for Wan2.2-5B and from 37.9\% to 43.9\% for LingBot-Video-30B. Physics-focused captions achieved the highest score for both backbones, reaching 33.2\% and 45.5\%, respectively. These results correspond to gains of 2.2 and 1.6 percentage points over raw captions and show that physics-focused captions outperform raw captions on the full dataset for both backbones.

\subsection{Image-Action-to-Video Evaluation}
\label{sec:ia2v_experiments}

\subsubsection{Implementation Details}
\label{sec:ia2v_implementation}

We train our IA2V model on AgiBot World-Beta~\cite{agibotworld2025} together with over 500 hours of general-domain T2V/TI2V data to preserve its original TI2V and general-domain generation capabilities. We evaluate the model on a fixed subset of 5K video clips from the AgiBotWorld-Alpha corpus~\cite{agibotworld2025}.
All evaluations use the final causal few-step student after the complete distillation pipeline. Unless otherwise specified, the student uses four denoising steps per chunk, with no explicit CFG at inference. At each control step, the input command is represented as the EE20 end-effector vector introduced in Section~\ref{sec:ee20_representation}.

\subsubsection{Performance Comparison}
\label{sec:performance}

For comparison, we evaluate DreamDojo-AgiBot-14B~\cite{dreamdojo} and A2World~\citep{a2world}, both of whose training corpora cover AgiBot robot platforms. Each model retains its native action representation. Specifically, DreamDojo uses a 22-dimensional action vector covering the arm joints, grippers, head, waist, and base velocity, whereas A2World adopts a 14-dimensional representation consisting of end-effector translation and rotation increments in the local coordinate frame. In contrast, our model relies solely on the EE20 end-effector target pose representation.

We evaluate standard fidelity metrics, including PSNR, SSIM, and LPIPS. In addition, Motion MAE measures the mean absolute error between the predicted and ground-truth adjacent-frame RGB differences.

Beyond overall video fidelity, we further evaluate whether the generated videos faithfully reflect the commanded robot actions in terms of trajectory consistency and the physical plausibility of robot-object interactions. Specifically, we report WorldArena-style metrics~\cite{shang2026worldarena}, including: (1) \textbf{Trajectory Accuracy} (Traj. Acc.), which measures the agreement between detected image-space gripper trajectories using dynamic time warping; (2) \textbf{Depth Accuracy} (Depth Acc.), which measures the consistency between depth maps estimated from the generated and ground-truth videos; and (3) \textbf{Interaction Quality}, a VLM-assigned score ranging from 1 to 5 that evaluates the physical plausibility of robot-object interactions, with higher scores indicating more plausible interactions.

As shown in Table~\ref{tab:ia2v_agibot_external}, StrucPhysVideo-IA2V consistently outperforms DreamDojo-AgiBot across all evaluated metrics. Beyond higher frame-level fidelity (PSNR, SSIM, and LPIPS) and lower motion error, our model achieves substantially higher trajectory accuracy (0.8330 vs. 0.7905), while also improving depth consistency and interaction quality. These results indicate that StrucPhysVideo-IA2V not only produces more visually faithful videos, but also more accurately reflects the commanded robot actions and the resulting physical interactions.

\begin{table}[H]
\centering
\small
\setlength{\tabcolsep}{4pt}
\begin{tabular}{lccccccc}
\toprule
Method           &PSNR $\uparrow$ &SSIM $\uparrow$ &LPIPS $\downarrow$ &Motion MAE $\downarrow$ &Traj. Acc. $\uparrow$ &Depth Acc. $\uparrow$ &Interaction $\uparrow$ \\
\midrule
DreamDojo-AgiBot &20.86           &0.8277          &0.1825             &0.03509                 &0.7905                &0.9736                &3.366\\
StrucPhysVideo-IA2V &22.12           &0.8365          &0.1602             &0.03368                 &0.8330                &0.9854                &3.470\\
\bottomrule
\end{tabular}
\caption{Quantitative performance on 5K video clips. Metrics exclude the reference frame. PSNR is in dB. Traj. Acc. and Depth Acc. are normalized to $[0,1]$, with higher values indicating better agreement.}
\label{tab:ia2v_agibot_external}
\end{table}

\begin{figure}[!tp]
  \centering
  \includegraphics[width=\linewidth,height=\textheight,keepaspectratio]{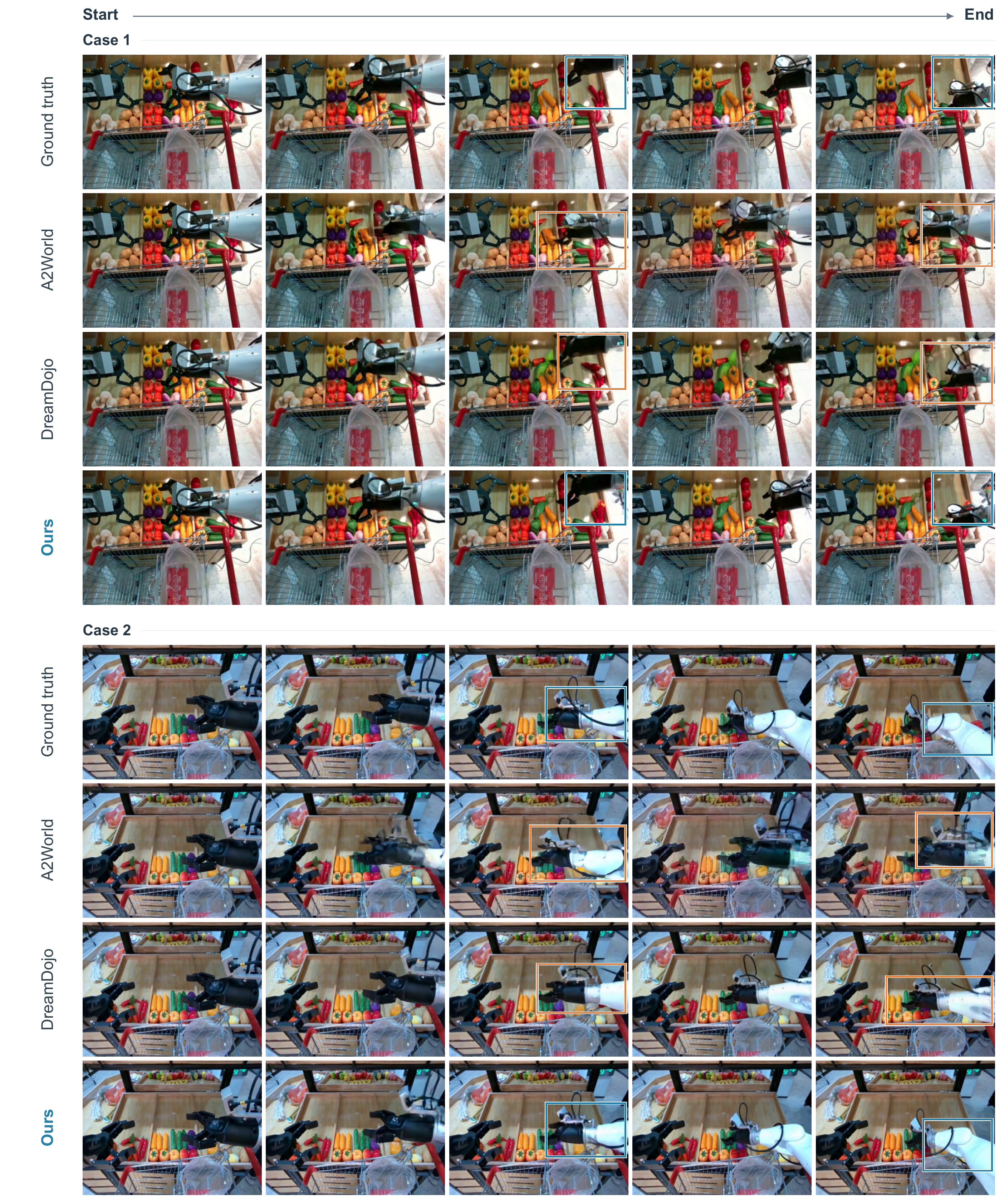}
  \caption{AgiBot head-view rollouts at 0, 1, 2, 3, and 4 seconds. Each group shows GT, A2World, DreamDojo-AgiBot, and ours. All start from the same head image. All three models produce arm motion, but the generated trajectories and local robot geometry differ from ground truth. A2World also exhibits visible appearance distortion in these examples.}
  \label{fig:ia2v_agibot_qualitative}
\end{figure}

To match A2World's default 21-frame, $256\times256$-per-view generation setting, we recompute all metrics on the same head-camera crop, resized to $256\times256$, over the first 21 frames at 5 fps; the results are shown in Table~\ref{tab:ia2v_agibot_three_models}. The corresponding qualitative rollouts are shown in Figure~\ref{fig:ia2v_agibot_qualitative}.

\begin{table}[H]
\centering
\small
\setlength{\tabcolsep}{4pt}
\begin{tabular}{lccccccc}
\toprule
Method           &PSNR $\uparrow$ &SSIM $\uparrow$ &LPIPS $\downarrow$ &Motion MAE $\downarrow$ &Traj. Acc. $\uparrow$ &Depth Acc. $\uparrow$ &Interaction $\uparrow$ \\
\midrule
A2World          &17.71           &0.7110          &0.2105             &0.04316                 &-                     &-                     &-    \\
DreamDojo-AgiBot &22.14           &0.8415          &0.1687             &0.03338                 &0.7853                &0.9685                &3.365 \\
StrucPhysVideo-IA2V &22.89           &0.8571          &0.1550             &0.03172                 &0.8287                &0.9827                &3.470 \\
\bottomrule
\end{tabular}
\caption{Quantitative comparison on $256\times256$ head-view grid and frames 1--20 after the reference frame. A2World additionally receives initial wrist views and observed-pose-derived controls. Qualitative comparison is provided in Figure~\ref{fig:ia2v_agibot_qualitative}.}
\label{tab:ia2v_agibot_three_models}
\end{table}

\subsubsection{Ablation on Teacher Dropout and Guidance}
\label{sec:ia2v_ablation}

We investigate how unconditional dropout during bidirectional action-injection training and teacher-side action CFG during distillation affect the final distilled student. We compare three teacher-training settings: (1) no dropout for either action or text conditions; (2) dropout applied only to the action condition; and (3) independent dropout applied to both action and text conditions. The CFG settings in Table~\ref{tab:robot_cfg_ablation} refer to the teacher-side, rather than to student inference. All reported metrics are measured using four-step, single-branch student rollouts without explicit CFG.

As shown in Table~\ref{tab:robot_cfg_ablation}, we make three interesting observations: (1) even without CFG, applying dropout only to the action condition does not degrade performance; (2) action CFG substantially improves both action conditioning and IA2V generation quality. In our setting, an action CFG scale of 2 performs best, while further increasing the scale becomes detrimental; and (3) even in IA2V mode, where text CFG is not used at inference time, introducing a small amount of text dropout during training still benefits IA2V performance.
Note that text dropout is applied only to the general-domain T2V/TI2V data mixed into training, while the IA2V data use no text prompts.

\begin{table}[H]
\centering
\small
\setlength{\tabcolsep}{4pt}
\resizebox{0.7\linewidth}{!}{
\begin{tabular}{clccc}
\toprule
Action dropout &Text dropout   &Teacher CFG setting   &Motion MAE $\downarrow$ &Traj. Acc. $\uparrow$ \\
\midrule
0\%            &0\%            &w/o CFG               &0.03537                 &0.7913 \\
\noalign{\vspace{1.5pt}}\hdashline\noalign{\vspace{1.5pt}}
15\%           &0\%            &w/o CFG               &0.03455                 &0.7896 \\
15\%           &0\%            &Action CFG $s_a=2$    &\underline{0.03449}     &0.8270 \\
15\%           &0\%            &Action CFG $s_a=3$    &0.03517                 &\textbf{0.8284} \\
15\%           &0\%            &Action CFG $s_a=5$    &0.03583                 &0.8195 \\
\noalign{\vspace{1.5pt}}\hdashline\noalign{\vspace{1.5pt}}
15\%           &5\%            &Action CFG $s_a=2$    &\textbf{0.03432}        &\underline{0.8272} \\
\bottomrule
\end{tabular}
}
\caption{Effects of teacher dropout and guidance on final distilled students. Dropout rates refer to bidirectional action-injection training. The reported action CFG scale is used both by the AR teacher to generate causal ODE distillation trajectories and by the bidirectional teacher to compute the real score during DMD. Bold and underlined values indicate the best and second-best values in each metric column, respectively.}
\label{tab:robot_cfg_ablation}
\end{table}

\section{Conclusion}
\label{sec:conclusion}
We presented \textbf{StrucPhysVideo}, a video world model developed through physically grounded data curation, annotation, and post-training. Our pipeline combines video sourcing and filtering with structured captions and physical phenomenon tags to preserve and describe how objects move, interact, and change state. By explicitly annotating participating objects, material properties, temporally localized interactions, and object motion distinct from camera movement, these data provide physically grounded supervision for adapting video foundation models toward physical world modeling. We further extend this framework to action-conditioned video prediction using paired action--video data, connecting robot commands with the resulting evolution of visual scenes.

StrucPhysVideo achieves 45.5\% on Physics-IQ Verified, while caption ablations across two video backbones consistently demonstrate the benefit of physics-focused language supervision over raw captions. Our action-conditioned experiments further provide an initial step toward interactive video world models that respond to embodied actions.

Looking forward, an important direction is to bridge world prediction and downstream embodied policies, where generated visual dynamics can support interaction, planning, and control in the physical world. This requires extending the current model to longer-horizon streaming rollouts while maintaining physical consistency over time, as well as further reducing inference latency---potentially to one- or two-step generation---to enable more responsive interaction. Beyond generation efficiency, supporting richer action spaces, and enabling cross-embodiment generalization across diverse robot platforms are also important steps toward more general embodied world models. 
Together, we hope StrucPhysVideo provides a foundation for advancing video generation from modeling observable physical dynamics toward interacting with the physical world.

\section{Contributors}
\label{sec:contributors}

All core contributors listed below contributed \textbf{equally} to this work and are listed \textbf{alphabetically by first name}.

\noindent\textbf{Data Infra \& Model Training:} Enhui Ma, Kaiwen Guo, Tingrui Zhang, Wei Song, Yingshui Tan.

\noindent\textbf{Project Leads:} Jianhua Xu, Tong Zhang.

\noindent\textbf{Project Sponsors:} Jianhua Xu, Kaicheng Yu.

\bibliographystyle{unsrtnat}
\bibliography{ref}

\clearpage
\appendix
\section{StrucPhysVideo-TI2V Training and Inference Algorithms}
\label{app:ti2v-algorithms}

Algorithm~\ref{alg:awomo-training} summarizes one stochastic training step for StrucPhysVideo-TI2V. The frozen encoders first construct the multimodal condition and normalized clean video latent. The procedure then samples a flow time and noise realization, forms the corrupted latent, and restores its first temporal position from the clean reference latent. The Transformer predicts the flow-matching velocity, while masking excludes the observed reference position from the denoising objective. Consequently, each update trains only the future latent trajectory while retaining the first frame as context.

\begin{algorithm}[H]
  \caption{StrucPhysVideo-TI2V Single-Step Training}
  \label{alg:awomo-training}
  \begin{algorithmic}[1]
    \Require Video $\mathbf{V}=\{I_0,I_1,\ldots,I_T\}$, caption $c$, frozen multimodal encoder $\mathcal{Q}$, frozen VAE encoder $\mathcal{E}$
    \State $\mathbf{h}_c\leftarrow\mathcal{Q}(c,I_0)$
    \State $\mathbf{z}_0\leftarrow\mathcal{N}(\mathcal{E}(\mathbf{V}))$
    \State Extract the reference latent $\mathbf{z}_0^{\mathrm{ref}}$ from the first temporal position of $\mathbf{z}_0$
    \State Sample $\boldsymbol{\epsilon}\sim\mathcal{N}(0,\mathbf{I})$ and a logit-normal flow time $\tau$
    \State $\mathbf{z}_\tau\leftarrow(1-\tau)\mathbf{z}_0+\tau\boldsymbol{\epsilon}$
    \State Replace the first temporal position of $\mathbf{z}_\tau$ with $\mathbf{z}_0^{\mathrm{ref}}$
    \State $\mathbf{v}_\theta\leftarrow\operatorname{StrucPhysVideo}(\mathbf{z}_\tau,\tau,\mathbf{h}_c)$
    \State $\mathbf{v}^{\star}\leftarrow\boldsymbol{\epsilon}-\mathbf{z}_0$
    \State Mask the reference position in $\mathbf{v}_\theta$ and $\mathbf{v}^{\star}$
    \State Compute the timestep-weighted mean-squared error and update the Transformer
  \end{algorithmic}
\end{algorithm}

Algorithm~\ref{alg:awomo-inference} describes the corresponding conditional sampling process. It initializes the future trajectory from Gaussian noise but fixes the first temporal position to the encoded reference image. At every solver step, conditional and negative-condition predictions are combined through classifier-free guidance before the Flow-UniPC update. Re-clamping the clean reference latent after each update prevents the observed frame from drifting as the future frames are generated. The converged latent is finally inverse-normalized and decoded into the output video.

\begin{algorithm}[H]
  \caption{StrucPhysVideo-TI2V Inference}
  \label{alg:awomo-inference}
  \begin{algorithmic}[1]
    \Require Caption $c$, reference image $I_0$, sampling steps $K$, text-guidance scale $s_t$
    \State $\mathbf{h}_c\leftarrow\mathcal{Q}(c,I_0)$ and $\mathbf{h}_u\leftarrow\mathcal{Q}(c_{\mathrm{neg}},I_0)$
    \State $\mathbf{z}_0^{\mathrm{ref}}\leftarrow\mathcal{N}(\mathcal{E}(I_0))$
    \State Sample an initial video latent $\mathbf{z}_K\sim\mathcal{N}(0,\mathbf{I})$
    \State Replace the first temporal position of $\mathbf{z}_K$ with $\mathbf{z}_0^{\mathrm{ref}}$
    \For{$k=K,K-1,\ldots,1$}
      \State Predict $\mathbf{v}_{\mathrm{cond}}$ from $(\mathbf{z}_k,\mathbf{h}_c)$ and $\mathbf{v}_{\mathrm{uncond}}$ from $(\mathbf{z}_k,\mathbf{h}_u)$
      \State $\widehat{\mathbf{v}}\leftarrow\mathbf{v}_{\mathrm{uncond}}+s_t(\mathbf{v}_{\mathrm{cond}}-\mathbf{v}_{\mathrm{uncond}})$
      \State $\mathbf{z}_{k-1}\leftarrow\operatorname{FlowUniPCStep}(\mathbf{z}_k,\widehat{\mathbf{v}},\tau_k)$
      \State Replace the first temporal position of $\mathbf{z}_{k-1}$ with $\mathbf{z}_0^{\mathrm{ref}}$
    \EndFor
    \State $\widehat{\mathbf{V}}\leftarrow\mathcal{D}(\mathcal{N}^{-1}(\mathbf{z}_0))$
    \State \Return $\widehat{\mathbf{V}}$
  \end{algorithmic}
\end{algorithm}

\clearpage
\section{Caption Example}
\label{app:caption-examples}

The two caption examples in this appendix describe the video shown as Case~1 in Figure~\ref{fig:ti2v-three-model-comparison}. They provide the real structured caption and the corresponding real raw caption for the same blue-ball and wooden-track sequence.

\subsection{Real Structured Caption Example}
\label{app:structured-caption-example}

Listing~\ref{lst:structured-caption-example} presents a real structured caption for a five-second clip in which a blue ball moves through a wooden marble run. Its four top-level keys correspond to the information groups defined in Section~\ref{sec:physical_caption}. \texttt{global\_scene\_dynamics} describes the setting and progression of the physical event. \texttt{camera\_motion\_and\_illumination} separates camera state, composition, and lighting from object motion. \texttt{world\_knowledge\_and\_material\_properties} records physical priors governing gravity-driven descent, track-constrained rolling, tangent-direction motion after leaving the track, and the spatial alignment required to enter the lower track. \texttt{entity\_level\_temporal\_behaviors} records the ball, wooden run, and human hand together with their visual attributes, spatial relationships, and timestamped behaviors.

For readability, Listing~\ref{lst:structured-caption-example} displays the caption as a formatted, multiline JSON object. The actual input to the StrucPhysVideo text encoder is a compact, single-line JSON string obtained by serializing the \texttt{caption} object, without inter-field spaces or indentation.

\begin{lstlisting}[
  basicstyle=\ttfamily\scriptsize,
  breaklines=true,
  breakatwhitespace=false,
  columns=fullflexible,
  keepspaces=true,
  showstringspaces=false,
  numbers=none,
  frame=single,
  caption={Formatted structured-caption JSON example.},
  label={lst:structured-caption-example}
]
{
  "global_scene_dynamics": "The video features a wooden marble run set up on a white table. The run consists of a long, wavy track on the left and a straight track on the right, both made of light-colored wood. A small blue ball is released at the start of the wavy track. As it rolls down, it follows the curves of the track, eventually reaching the end of the straight section. When the ball leaves the end of this track, it must continue straight along the track's outgoing tangent direction, with no sideways deflection or skew, then fall naturally under gravity into a second wooden track positioned below the tabletop. The lighting is bright and even, creating a clean and focused atmosphere on the mechanical action of the ball.",
  "camera_motion_and_illumination": {
    "camera_motion": "The camera is stationary throughout the entire video, providing a fixed, high-angle perspective of the marble run and the table.",
    "dominant_color": "Red",
    "shot_scale": "Wide",
    "viewing_angle": "High angle",
    "lens_scale": "Medium",
    "frame_composition": "Left heavy",
    "illumination_quality": "Soft light",
    "illumination_source": "Artificial light"
  },
  "world_knowledge_and_material_properties": [
    "Gravity drives the ball along descending sections of the wooden track.",
    "While constrained by the track, the rolling ball follows the track's curved path.",
    "When the ball leaves the end of the track, inertia carries it forward along the outgoing tangent direction rather than causing a sideways deflection.",
    "Once unsupported, the ball follows a gravity-driven downward trajectory, and entering the lower track requires the two track sections to be spatially aligned."
  ],
  "entity_level_temporal_behaviors": [
    {
      "entity_name": "blue ball",
      "visual_description": "A small, solid blue spherical object used as a marble in the run.",
      "timestamped_behaviors": [
        {
          "time_interval": "[0.0s - 0.2s]",
          "behavior": "is dropped into the starting hole of the wooden track"
        },
        {
          "time_interval": "[0.2s - 2.5s]",
          "behavior": "rolls down the wavy wooden track"
        },
        {
          "time_interval": "[2.5s - 4.0s]",
          "behavior": "rolls along the straight wooden track"
        },
        {
          "time_interval": "[4.0s - 5.0s]",
          "behavior": "leaves the end of the straight track in its outgoing tangent direction without sideways deflection, then falls under gravity into the lower wooden track below the tabletop"
        }
      ],
      "spatial_location": "starts at the top left, moves through the center, and exits at the bottom right",
      "relative_scale": "small",
      "form_and_color": "spherical and blue",
      "surface_properties": "smooth",
      "appearance_attributes": "solid blue color with no visible markings",
      "scene_relationship": "rolls along the wooden tracks and is released by a hand",
      "spatial_orientation": "rolling",
      "body_pose": "",
      "facial_expression": "",
      "apparel": "",
      "perceived_gender": "",
      "visible_skin_attributes": ""
    },
    {
      "entity_name": "wooden marble run",
      "visual_description": "A set of light-colored wooden tracks designed for a marble to roll through.",
      "timestamped_behaviors": [
        {
          "time_interval": "[0.0s - 5.0s]",
          "behavior": ""
        }
      ],
      "spatial_location": "occupies the left and center portions of the frame",
      "relative_scale": "large",
      "form_and_color": "long, wavy and straight tracks in light brown wood",
      "surface_properties": "smooth wood grain",
      "appearance_attributes": "features a wavy section on the left and a straight section on the right",
      "scene_relationship": "serves as the path for the blue ball",
      "spatial_orientation": "horizontal and slightly diagonal",
      "body_pose": "",
      "facial_expression": "",
      "apparel": "",
      "perceived_gender": "",
      "visible_skin_attributes": ""
    },
    {
      "entity_name": "human hand",
      "visual_description": "A person's hand visible at the start of the video.",
      "timestamped_behaviors": [
        {
          "time_interval": "[0.0s - 0.2s]",
          "behavior": "drops the blue ball into the track"
        }
      ],
      "spatial_location": "top left corner of the frame",
      "relative_scale": "medium",
      "form_and_color": "natural skin tone",
      "surface_properties": "smooth",
      "appearance_attributes": "only the fingers and part of the palm are visible",
      "scene_relationship": "releases the blue ball into the track",
      "spatial_orientation": "reaching down",
      "body_pose": "fingers pinching the ball",
      "facial_expression": "",
      "apparel": "",
      "perceived_gender": "male",
      "visible_skin_attributes": "light skin tone"
    }
  ]
}
\end{lstlisting}

\clearpage
\subsection{Real Raw Caption Example}
\label{app:raw-caption-example}

Listing~\ref{lst:raw-caption-example} provides the corresponding real raw caption before conversion into the structured representation above. It describes the same blue-ball and wooden-track scene as continuous prose, including the intended motion, physical constraints, stationary scene elements, camera requirements, and visual style.

\begin{lstlisting}[
  basicstyle=\ttfamily\scriptsize,
  breaklines=true,
  breakatwhitespace=true,
  columns=fullflexible,
  keepspaces=false,
  showstringspaces=false,
  numbers=none,
  frame=single,
  caption={Real raw-caption example corresponding to Listing~\ref{lst:structured-caption-example}.},
  label={lst:raw-caption-example}
]
A realistic physics simulation featuring exactly one single small blue ball rolling down a wooden marble track. The same single blue ball must remain continuously visible and physically consistent throughout the entire sequence. The ball starts at the upper part of the inclined wooden track and moves continuously under gravity. The action starts with the ball dropping directly down into the top hole of the starting cylindrical wooden block. It rolls smoothly along the carved wavy path, following every curve and turn of the track while maintaining constant contact with the wooden surface. The ball should rotate as it rolls, with realistic friction and inertia, without sliding, bouncing, floating, or passing through the track.

All wooden tracks, the tabletop, and every background object remain completely stationary, rigid, and unchanged throughout the entire video. The movement is continuous and stable, with natural momentum conservation and realistic interaction between the ball and the wooden surface.

Maintain the original camera viewpoint and real-world appearance. Preserve the wooden texture, lighting, shadows, and scale of the scene. The animation should look like a real physical experiment recorded with a camera, with accurate gravity, collision, and rolling dynamics. No deformation of objects, no extra objects, no movement of the background tracks, no unrealistic motion.

Style: realistic laboratory physics demonstration, high-quality slow-motion video, natural lighting, real-world physics, accurate object interaction.
\end{lstlisting}

\end{document}